\documentclass[article]{Definitions/preprint} 

\usepackage{epstopdf}
\usepackage{caption}
\usepackage{subcaption}
\usepackage{dsfont}
\usepackage{xurl}
\usepackage{rotating}
\usepackage{longtable}
\usepackage{tabularx}
\usepackage[normalem]{ulem} 
\def\UrlBreaks{\do\/\do_}

\usepackage{amssymb}
\usepackage{pifont}

\DeclareMathOperator*{\argmax}{arg\,max}

\title{Unraveling Anomalies in Time: Unsupervised Discovery and Isolation of Anomalous Behavior in Bio-regenerative Life Support System Telemetry}

\author{ 
   \href{https://orcid.org/0000-0003-2264-9495}{\includegraphics[scale=0.06]{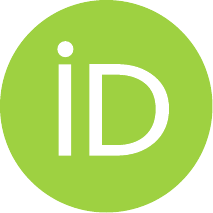}\hspace{1mm}Ferdinand Rewicki}
        \\
    Institute of Data Science\\
	German Aerospace Center\\
	07745 Jena, Germany \\
	\texttt{ferdinand.rewicki@dlr.de} \\
    \And
	\href{https://orcid.org/0000-0002-2219-3590}{\includegraphics[scale=0.06]{Definitions/logo-orcid.pdf}\hspace{1mm}Jakob Gawlikowski} \\
	Institute of Data Science\\
	German Aerospace Center\\
	07745 Jena, Germany \\
	\texttt{jakob Gawlikowski@dlr.de} \\
	\And
	\href{https://orcid.org/0000-0001-5413-2234}{\includegraphics[scale=0.06]{Definitions/logo-orcid.pdf}\hspace{1mm}Julia Niebling} \\
	Institute of Data Science\\
	German Aerospace Center\\
	07745 Jena, Germany \\
	\texttt{julia.niebling@dlr.de} \\
    \And
	\href{https://orcid.org/0000-0002-3193-3300}{\includegraphics[scale=0.06]{Definitions/logo-orcid.pdf}\hspace{1mm}Joachim Denzler} \\
	Institute of Computer Science\\
	Friedrich-Schiller University\\
	07743 Jena, Germany \\
	\texttt{joachim.denzler@uni-jena.de} \\
}

\renewcommand{\shorttitle}{Unraveling Anomalies in Time}

\hypersetup{
pdftitle={Unraveling Anomalies in Time: Unsupervised Discovery and Isolation of Anomalous Behavior in Bio-regenerative Life Support System Telemetry},
pdfsubject={machine-learning,anomaly-detection,time-series},
pdfauthor={Ferdinand Rewicki, Joachim Denzler, Julia Niebling},
pdfkeywords={Anomaly Detection, Time Series, Machine Learning, Deep Learning, Benchmark},
}

\begin{document}
\maketitle

\begin{abstract}
The detection of abnormal or critical system states is essential in condition monitoring. 
While much attention is given to promptly identifying anomalies, a retrospective analysis of these anomalies can significantly enhance our comprehension of the underlying causes of observed undesired behavior. 
This aspect becomes particularly critical when the monitored system is deployed in a vital environment.
In this study, we delve into anomalies within the domain of Bio-Regenerative Life Support Systems (BLSS) for space exploration. 
We analyze anomalies found in telemetry data stemming from the EDEN ISS space greenhouse in Antarctica, using MDI and DAMP, two glassbox methods for anomaly detection based on density estimation and discord discovery respectively.
We employ time series clustering on anomaly detection results to categorize various types of anomalies in both uni- and multivariate settings. 
We then assess the effectiveness of these methods in identifying systematic anomalous behavior.
Additionally, we illustrate that the anomaly detection methods MDI and DAMP produce complementary results, as previously indicated by research.
\end{abstract}


\section{Introduction}
Bio-regenerative Life Support Systems (BLSSs) are artificial ecosystems that consist of multiple symbiotic relationships.
BLSSs are crucial for sustaining long-duration space missions by facilitating food production and managing essential material cycles for respiratory air, water, biomass, and waste. 
The EDEN NEXT GEN Project, part of the EDEN roadmap at the German Aerospace Center (DLR), aims to develop a fully integrated ground demonstrator of a BLSS comprising all subsystems, with the ultimate goal of realizing a flight-ready BLSS within the next decade.
This initiative builds upon insights from the EDEN ISS project, which investigated controlled environment agriculture (CEA) technologies for space exploration. 
EDEN ISS, a near-closed-loop research greenhouse deployed in Antarctica from 2017 to 2021, focused on crop production, including lettuces, bell peppers, leafy greens, and various herbs.
To ensure the safe and stable operation of BLSSs, we explore methods to mitigate risks regarding system health, particularly regarding food production and nourishment shortages for isolated crews. 
Given the absence of clear definitions for unhealthy system states and the unavailability of ground truth data, we investigate unsupervised anomaly detection methods.
Unsupervised anomaly detection targets deviations or irregularities from expected or standard behavior in the absence of labelled training data.
Choosing the appropriate method from the plethora of available options is challenging due to differing strengths in detecting certain types of anomalies, as no universal method exists \cite{Laptev2015}. 

To address this challenge, we conducted a comparative analysis of six unsupervised anomaly detection methods, differing in complexity in \cite{Rewicki2023}.
Three of these methods are classical machine learning techniques, while the remaining three are based on deep learning.
The primary questions in this comparison have been: (1) "Is it worthwhile to sacrifice the interpretability of classical methods for potentially superior performance of deep learning methods?" and (2) "What different types of anomalies are the methods capable of detecting?"
The findings underscored the efficacy of two classical methods, Maximally Divergent Intervals (MDI) \cite{Barz2018} and MERLIN \cite{Nakamura2020}, which not only performed best individually but also complement each other in terms of the detected anomaly types.

\begin{figure} 
    \centering 
    \includegraphics[width=.9\columnwidth]{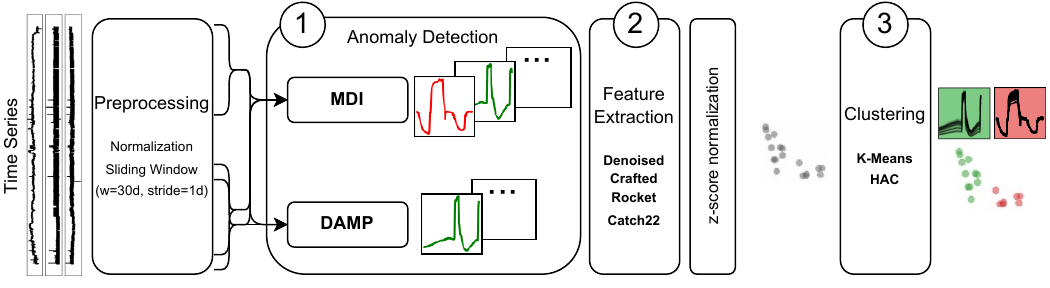} 
    \caption{Overview of our approach to derive different types of anomalous behavior from unlabelled time series. \textcircled{\footnotesize 1} MDI and DAMP are applied to time series data to obtain anomalous sequences. \textcircled{\footnotesize 2} Features are extracted from sequences and  \textcircled{\footnotesize 3} clustering is applied to derive different anomaly types, marked by color.}
    \label{fig:method:pipeline}
\end{figure}

Building upon our findings in \cite{Rewicki2023}, we analyze telemetry data\footnote{The code to reproduce our results is available at \href{https://gitlab.com/dlr-dw/unraveling-anomalies-in-time/-/tree/v1.0.0}{https://gitlab.com/dlr-dw/unraveling-anomalies-in-time/-/tree/v1.0.0}.} from the EDEN ISS subsystems for the mission year 2020. 
Our objectives include discovering anomalous behavior, differentiating different types of anomalies, and identifying recurring anomalous behavior. Figure \ref{fig:method:pipeline} outlines our approach.
We apply MDI and Discord Aware Matrix Profile (DAMP), another algorithm for Discord Discovery similar to MERLIN, for anomaly detection, to obtain univariate and multivariate anomalies and extract four sets of features from the anomalous sequences. We finally apply K-Means and Hierachical Agglomerative Clustering (HAC) to obtain clusters representing similar anomalous behavior. Experimental validation addresses five research questions related to the complementarity of MDI and DAMP (RQ1), optimal feature selection (RQ2), superior clustering algorithms (RQ3), types of isolated anomalies (RQ4), and identification of recurring abnormal behavior (RQ5).

\section{Related Work}
\label{sec:relatedwork}

While the field of anomaly detection has witnessed great interest and an enormous number of publications in recent years, particularly in endeavors focusing on timely anomaly detection.
Scant attention has been directed towards the critical task of categorizing and delineating various forms of anomalous behavior. 
Sohn et al. (2023) discriminate anomalous images into clusters of coherent anomaly types using bag-of-patch-embeddings representations and HAC with Ward linkage \cite{Sohn2023}. 
In \cite{Rewicki2023} we evaluated six anomaly detection methods with varying complexity regarding their ability to detect certain shape-based anomaly types in univariate time series data. We showed, that the density estimation-based method Maximally Divergent Intervals (MDI) \cite{Barz2018} and the Discord Discovery method MERLIN \cite{Nakamura2020} not only deliver the best individual results in this study but yield complementary results in terms of the types of anomalies they can detect.
Tafazoli et al. (2023) recently proposed a combination of the Matrix profile, which is the underlying technique behind discord discovery, with the "Canonical Time-series Chracteristics" (Catch22) features, that we employ as one of our methods for feature extraction \cite{Tafazoli2023}.
Ruiz et al. (2021) experimentally compared algorithms for time series classification, and found that "the real winner of this experimental analysis is ROCKET [Author's note: Random Convolutional Kernel Transform]"\cite{Ruiz2021} as it is the best ranked and by far the fasted classifier in their study \cite{Ruiz2021} . We use ROCKET as one method to derive features from time series.
Anomaly detection has also garnered interest in the CEA domain. 
While the works \cite{Adkinson2021,AraujoZanella2022,Joaquim2022} proposed various methods for anomaly detection in the CEA and Smart Farming domain, other studies have explored anomaly detection's utility in enhancing greenhouse control \cite{Canadas2017} or monitoring plant growth \cite{Choi2021,Xhimitiku2021}.
However, there has been limited effort to extract potential systematic behavior from anomaly detection results. This work aims to address this gap by contributing towards the derivation of systematic behaviors from anomaly detection outcomes in the telemetry data of the EDEN ISS space greenhouse.

\section{Methodology}
\label{sec:method}
In the following section we introduce our pipeline outlined in Figure \ref{fig:method:pipeline} to derive different types of anomalous behavior from unlabeled time series.
We start by defining time series data and subsequences, followed by the introduction of the methods for anomaly detection and feature extraction. 
Finally, we present the measures  we use to evaluate the quality of clustering results.
While we evaluate our approach purely based on our specific use case, its generic nature allows it to be applied to different time series mining tasks.

Time series are sequential data that are naturally ordered by time. 
We define a regular time series as an ordered set of observations made at equidistant intervals based on \cite{Nakamura2020}:
\begin{Definition}
The regular \textbf{time series} $\mathcal{T}$ with length $N \in \mathbb{N}$ is defined as the set of pairs $\mathcal{T} = \lbrace{(t_n, \mathbf{x_n}) | t_n \leq t_{n+1}, 0 \leq n \leq N-1, t_{n+1} - t_n = c \rbrace}$ with $\mathbf{x_n} \in \mathbb{R}^D$ being the data points having $D$ behavioural attributes and $t_n \in \mathbb{N}, n \leq N$ the equidistant timestamps the data refer to. 
For $D=1$, $\mathcal{T}$ is called univariate, and for $D>1$, $\mathcal T$ is called multivariate.
\end{Definition}

As time series are usually not analyzed en bloc, we define a subsequence as a contiguous subset of the time series: 
\begin{Definition}
The \textbf{subsequence} $\mathcal{S}_{a,b} \subseteq \mathcal{T}$ of the times series $\mathcal{T}$, with length $L = b-a+1 > 0$ is given by $\mathcal{S}_{a,b} := \lbrace{ (t_n, \mathbf{x_n}) | 0 \leq a \leq n \leq b < N \rbrace }$. 
For multivariate time series $\mathcal{T}$, $\mathcal{S}^{(i)}_{a,b}$ with $i \in \mathbb{N}$ refers to the subsequence $\mathcal{S}_{a,b}$ in dimension $1 \leq i \leq D$
For brevity, we often omit the indices and refer to arbitrary subsequences as $\mathcal{S}$.
\end{Definition}

\subsection{Anomaly Detection}
In the following, we understand anomalies as collective anomalies, i.e. special subsequences $S$, that deviate notably from an underlying concept of normality. We focus on collective anomalies as we are interested in prolonged environmental issues that can significantly impact plant health and operational efficiency. We selected MDI and DAMP as they yielded not only the best individual but also complementary results in \cite{Rewicki2023}.

\textbf{MDI} \cite{Barz2018} is a density-based method for offline anomaly detection in multivariate, spatiotemporal data. 
We focus on temporal data in this study, providing definitions pertinent to this context. For comprehensive definitions, including spatial attributes, refer to \cite{Barz2018}.
MDI identifies anomalous subsequences in a multivariate time series $\mathcal{T}$ by comparing the probability density $p_\mathcal{S}$ of a subsequence $\mathcal{S} \subseteq \mathcal{T}$ to the density $p_\Omega$ of the remaining part $\Omega(\mathcal{S}) := \mathcal{T} \setminus \mathcal{S}$. 
These distributions are modeled using Kernel Density Estimation or Multivariate Gaussians. 
MDI quantifies the degree of deviation $\mathcal{D}(p_\mathcal{S}, p_\Omega)$ an unbiased variant of the Kullback-Leibler divergence.
The most anomalous subsequence $\mathcal{\tilde{S}}$ is identified by solving the optimization problem: $\mathcal{\tilde{S}} := \argmax_{\mathcal{S} \subseteq \mathcal{T}} \mathcal{D}(p_\mathcal{S}, p_{\Omega(\mathcal{S})})$.
MDI locates this most anomalous subsequence $\mathcal{\tilde{S}}$ by scanning all subsequences $\mathcal{S} \subseteq \mathcal{T}$ with lengths between predefined parameters $L_{min}, L_{max} \in \mathbb{N}$ and estimating the divergence $\mathcal{D}(p_\mathcal{S}, p_{\Omega(\mathcal{S})})$, which serves as the anomaly score.
The anomalous subsequences are selected by ranking all subsequences based on their anomaly score and applying Non-maximum suppression.
To adapt to large-scale data, MDI employs an interval proposal technique based on Hotelling’s $T^2$ method \cite{MacGregor1994}. 
This technique selects interesting subsequences based on point-wise anomaly scores rather than conducting full scans over the entire time series, motivated by the rarity of anomalies in time series by definition \cite{Barz2018}.
We set $L_{min}$ and $L_{max}$ to $144$ ($0.5$ days) and $288$ ($1$ day) empirically.

\textbf{DAMP} \cite{Lu2022} is a method for both offline and "effectively online" anomaly detection by discord discovery. 
The term "effectively online" was introduced by \cite{Lu2022} to classify algorithms that are not strictly online but where "the lag in reporting a condition has little or no impact on the actionability of the reported information" \cite{Lu2022}.
Given a subsequence $\mathcal{S}_{a_1,b_1}$ and a matching subsequence $\mathcal{S}_{a_2,b_2}$ with $b_1-a_1 = b_2-a_2 = L$, $\mathcal{S}_{a_1,b_1}$ is a \textit{non-self match} to $S_{a_2,b_2}$ with distance $d_{a_1,a_2}$ if $|a_1-a_2| \geq L$. $dist(\cdot,\cdot)$ denotes the z-normalized Euclidean distance.
The discord $\mathcal{\tilde{S}}$ of a time series $\mathcal{T}$ is defined as the subsequence with the maximum distance $d(\mathcal{\tilde{S}}, M_{\mathcal{\tilde{S}}})$ from its nearest non-self match $\mathcal{M_{\mathcal{\tilde{S}}}}$. 
To ascertain the discord of a time series, DAMP approximates the left matrix profile $P^{L(\mathcal{T})}$, a vector storing the z-normalized Euclidean distance between each subsequence of $\mathcal{T}$ and its nearest non-self match occurring before that subsequence.
DAMP comprises a forward and a backward pass. 
In the backward pass, each subsequence is assessed to determine if it constitutes the discord of the time series. 
Meanwhile, the forward pass aids in pruning data points that do not qualify as discord based on the best-so-far discord distance.
We set $L$ to $288$ ($1$ day) empirically.

\subsection{Feature Extraction}
The objective of our analysis is to identify particular anomaly types specific to the EDEN ISS telemetry dataset. Given the exploratory nature of this analysis, we examine four distinct feature extraction methods, which we refer to below as "feature sets" and elaborate on in this section.

\textbf{Denoised Subsequences} As a first feature set, we utilize the raw subsequences identified as anomalous by MDI or DAMP. 
These subsequences vary in length from $18$ to $447$ data points\footnote{We excluded anomalies with a length of fewer than five data points from our analysis.} in the univariate and from $13$ to $289$ in the multivariate case.
To compare sequences of differing lengths, we employ Dynamic Time Warping (DTW)  \cite{Berndt1994}. 
To enhance comparability, we apply moving average smoothing with a window size of five data points to eliminate high frequencies. 
The window size has been set empirically.
In subsequent discussions, we refer to this feature set as \texttt{Denoised}.

\textbf{Handcrafted Feature-Vectors} For the second feature set, we derive a nine-dimensional vector comprising simple statistical and shape-specific features. This vector encompasses the first four moments, i.e. mean, variance, kurtosis, and skewness, alongside the sequence length, the minimum and maximum values, and the positions of the minimum and maximum within the sequence. Following the computation of these feature vectors, we employ z-score normalization to standardize the features to a zero mean and unit standard deviation. In the following, we will refer to this feature set as  \texttt{Crafted}.

\textbf{Random Convolutional Kernel Transform (ROCKET)} \cite{Dempster2020} generates features from time series using a large number of random convolutional kernels. 
Each kernel is applied to every subsequence, yielding two aggregate features: maximum value (similar to global max pooling) and proportion of positive values (PPV) \cite{Dempster2020,Ruiz2021}. 
Pooling, akin to convolutional neural networks, reduces dimensionality and achieves temporal or spatial invariance, while PPV captures kernel correlation.
ROCKET employs 10,000 kernels with lengths $l \in \lbrace{7,9,11 \rbrace}$ and weights $\mathbf{w} \in \mathbb{R}^l$ sampled from the standard normal distribution $\mathbf{w} \sim \mathcal{N}(\mathbf{0},\mathbf{I})$. 
We apply Principal Component Analysis (PCA) to the z-normalized transformation outcome and use the first 10 components, also z-normalized, as final features to mitigate dimensionality issues. 
We reduced the number of kernels to 1000, finding no significant alteration in results. We refer to this feature set as \texttt{Rocket}.

\textbf{Canonical Time Series Characteristics (catch22)} \cite{Lubba2019} comprise 22 time series features derived from an extensive search through 4,791 candidates and 147,000 diverse datasets. 
These features, tailored for time series data mining, demonstrate strong classification performance and minimal redundancy \cite{Lubba2019}. 
They encompass various aspects such as distribution of values, temporal statistics, autocorrelation (linear and non-linear), successive differences, and fluctuation. 
We apply these features to each subsequence, resulting in a 22-dimensional feature vector per sequence. 
Features with a normalized variance exceeding a threshold (set empirically at 0.01) are selected and the chosen features are then z-normalized. 
We will refer to this feature set as \texttt{Catch22}.

\subsection{Time Series Clustering}

Time series clustering involves partitioning a dataset $\mathfrak{D}$ containing time series $\mathcal{T}^{(1)}, \dots, \mathcal{T}^{(|\mathfrak{D}|)}$ into $K$ disjoint subsets ${C}_k, k = 2, \dots,  K$, where each subset contains similar time series.
Similarity is measured using distance measures like Euclidean Distance or Dynamic Time Warping (DTW) \cite{Berndt1994}. In this study, we compare K-Means clustering \cite{MacQueen1967} with Hierarchical Agglomerative Clustering (HAC) to identify clusters of similar anomalous subsequences.

\textbf{K-Means} clustering \cite{MacQueen1967} partitions a set of $n$ observations into $K$ clusters by assigning each observation to the cluster with the nearest mean, minimizing within-cluster variance. 
The centroids $\mu_k, k = 1, \dots, K$ serve as cluster prototypes. 
However, vanilla K-Means may not yield optimal results as it randomly selects initial centroids, making it sensitive to seeding. 
To address this, \cite{Arthur2007} proposed K-Means$++$, which selects centroids with probabilities proportional to their contribution to the overall potential. 
K-Means$++$ is now a standard initialization strategy for K-Means clustering, including in our experiments.

\textbf{HAC} partitions a set of $n$ observations into a hierarchical structure of clusters. 
It begins by treating each data point as a separate cluster and then merges the closest clusters iteratively. 
The choice of a linkage criterion, determining the dissimilarity measure between clusters, is crucial. 
In our experiments, we adopt the \textit{Unweighted Pair-Group Method of Centroids (UPGMC)} linkage, which calculates the distance between clusters based on the distance between their centroids. 
Other common linkage criteria include \textit{Single}, \textit{Complete} and \textit{Ward} linkage, which respectively use the minimum (Single) or maximum (Complete) distance between points from different clusters as linkage criterion or minimize the within cluster variance (Ward).

\subsection{Quality Measures}

The \textbf{Silhouette Score (SSC)} \cite{Rousseeuw1987} is the standard measure for evaluating clustering results and quantifies both cohesion and separation within clusters. 
It is calculated by averaging over the Silhouette Coefficients $SSC_C$ for each cluster $C$, defined as:
\begin{equation} 
\label{eq:ssc} 
  SSC_C = \frac{1}{|C|} \sum_{\mathcal{S} \in C} \frac{idist(\mathcal{S}) - wdist(\mathcal{S})}{\max(wdist(\mathcal{S}), idist(\mathcal{S}))} \text{   .}
\end{equation}
Here, $wdist(\mathcal{S})$ represents the mean distance of object $\mathcal{S} \in C$ to all other elements within its own cluster $C$ (within-cluster distance), while $idist(\mathcal{S})$ denotes the smallest mean distance to elements in another cluster (inter-cluster distance). \cite{Rousseeuw1987}
SSC ranges from $-1$ to $1$, where $1$ indicates well-separated clusters, $0$ suggests overlapping clusters, and $-1$ implies misclassification of objects.
\\\\

To evaluate the quality of clustering outcomes, we introduce the \textbf{Synchronized Anomaly Agreement Index (SAAI)}.
Let
\begin{equation}
  A = \{\mathcal{S}^{(i)}_{a,b} | i, a, b \in \mathbb{N}, i \leq D, a < b, s(\mathcal{S}^{(i)}_{a,b}) > th\}
\end{equation}
be the set of univariate anomalies in the time series $\mathcal{T} = \{\mathcal{T}^{(1)}, ..., \mathcal{T}^{(D)}\}$ where $s(\cdot): \{\mathcal{S} | \mathcal{S} \subseteq \mathcal{T}\} \rightarrow [0,1]$ denotes a anomaly score function, and $th \in [0,1]$ represents the threshold for labeling a subsequence anomalous.
Furthermore, let
\begin{equation}
  A_S = \{ (\mathcal{S}^{(i)}_{a_i,b_i}, \mathcal{S}^{(j)}_{a_j,b_j)} | \mathcal{S}^{(i)}_{a_i,b_i}, \mathcal{S}^{(j)}_{a_j,b_j} \in A,  i < j, iou([a_i,b_i], [a_j,b_j]) > th_{iou} \}
\end{equation}
be the set of synchronized, i.e. temporally aligned, anomalies with $iou([a_i,b_i],\allowbreak[a_j,b_j])$ representing the time-interval \textit{Intersection over Union} of two subsequences $\mathcal{S}^{(i)}_{a_i,b_i}$ and $\mathcal{S}^{(j)}_{a_j,b_j}$. 
The threshold parameter $th_{iou} \in [0,1]$ determines the degree of temporal alignment.
Additionally, let 
\begin{equation}
  A^*_S \subseteq A_S
\end{equation}
denote the set of temporally aligned anomalies assigned to the same cluster, where $c(S^{(i)}_{a_i,b_i}) = c(S^{(j)}_{a_j,b_j})$, with $c(\mathcal{S}^{(i)}_{a_i,b_i})$ indicating the cluster of subsequence $\mathcal{S}^{(i)}_{a_i,b_i}$.
The SAAI of a clustering solution of univariate anomalous subsequences in the set of time series $\mathcal{T} = \{\mathcal{T}^{(1)}, ..., \mathcal{T}^{(D)}\}$ is defined as:

\begin{equation}
\label{eq:saai}
    SAAI := \lambda \frac{|A^*_S|}{|A_S|} - (1-\lambda)(\frac{1}{K} + \frac{n_\mathds{1}}{K}) + (1-\lambda) \text{ , } \lambda \in [0,1] \text{  . } 
\end{equation}

Here, the first term $\frac{|A^*_S|}{|A_S|}$ evaluates the ratio of temporally aligned anomalies in the same cluster among all temporally aligned anomalies. 
The second term serves as regularization, accounting for small cluster sizes ($\frac{1}{K}$) and clusters containing only a single anomaly ($\frac{n_\mathds{1}}{K}$), where $n_\mathds{1}$ represents the number of single-element clusters. The parameter $\lambda$ allows adjusting the influence of the penalty term.
$(1-\lambda)$ is added to scale the value of SAAI to the interval $[0,1]$, enabling the comparison of SAAI values with different weights $\lambda$.
In our experiments detailed in Section \ref{sec:results}, we set $\lambda = 0.5$.

The rationale behind this measure is, although we lack knowledge of the real anomaly clusters, we hypothesize that temporally aligned anomalies in similar measurements - such as those from the same sensor types - should cluster together, as they likely represent the same anomaly.
Higher values indicate better clustering solutions. In the supplementary materials~\ref{appendix:saai_intuition} we provide examples and further information to interpret SAAI.
\\\\
The \textbf{Gini-Index}\cite{Sitthiyot2020} is a metric for statistical dispersion or imbalance. 
Given a set of discrete values $X = {x_1, x_2, \dots, x_K}$, it is defined as:
\begin{equation} 
\label{eq:gini} 
G = \frac{\sum_{i=1}^K \sum_{j=1}^K |x_i-x_j|}{2n\sum_{i=1}^K x_i} \text{   .}
\end{equation}
The Gini-Index ranges from 0 to 1, with lower values indicating more equal distribution and higher values suggesting greater inequality. 
We use the Gini-Index to evaluate cluster size imbalance across various clusterings by applying Equation \ref{eq:gini} to the cluster sizes $x_1=|C_1|, \dots, x_K=|C_K|$ of each solution.

\section{Experimental Results}
\label{sec:results}
The experimental results were obtained by first applying MDI and DAMP in the uni- and multivariate case. Data from one subsystem is represented by one multivariate time series. 
We extract features from the detected anomalous subsequences and cluster them with number of clusters $2 \leq K \leq 20$. 
All experiments were run on an Intel Xeon Platinum 8260 CPU with 20GB of allocated memory
Table \ref{tab:hyperparams} in the supplemental materials lists the hyperparameter settings we used for our experiments.

\subsection{Dataset}
\label{results:dataset}
The \textit{edeniss2020} dataset \cite{Rewicki2024a} comprises equidistant sensor readings from 97 variables derived from the four subsystems of the EDEN ISS greenhouse, namely the Atmosphere Management System (AMS), Nutrient Delivery System (NDS), Illumination Control System (ICS) and Thermal Control System (TCS).
Our analysis focuses on data from the year 2020, representing EDEN ISS's third operational year. Table \ref{tab:edeniss_dataset} in the supplemental materials~\ref{appendix:dataset_subsystems} lists the measurements per subsystem.
The data is captured at a sampling rate of $1/300$ Hz and covers the range from 01/01/2020 - 12/30/2020. Every of the 97 univariate time series has a length of 105120 data points.
The readings from the AMS pertain to air properties in the greenhouse and service section, while those from the NDS relate to nutrient solution tanks and pressure measurements in the pipes connecting tanks and growth trays. 
ICS temperature readings are taken above each growth tray and beneath the corresponding LED lamps.

\subsection{RQ1: Are the results of MDI and DAMP complementary?}
\label{results:rq1}
\begin{figure}[t]
    \centering
    \begin{subfigure}[b]{0.45\textwidth}
        \centering
        \includegraphics[width=\textwidth]{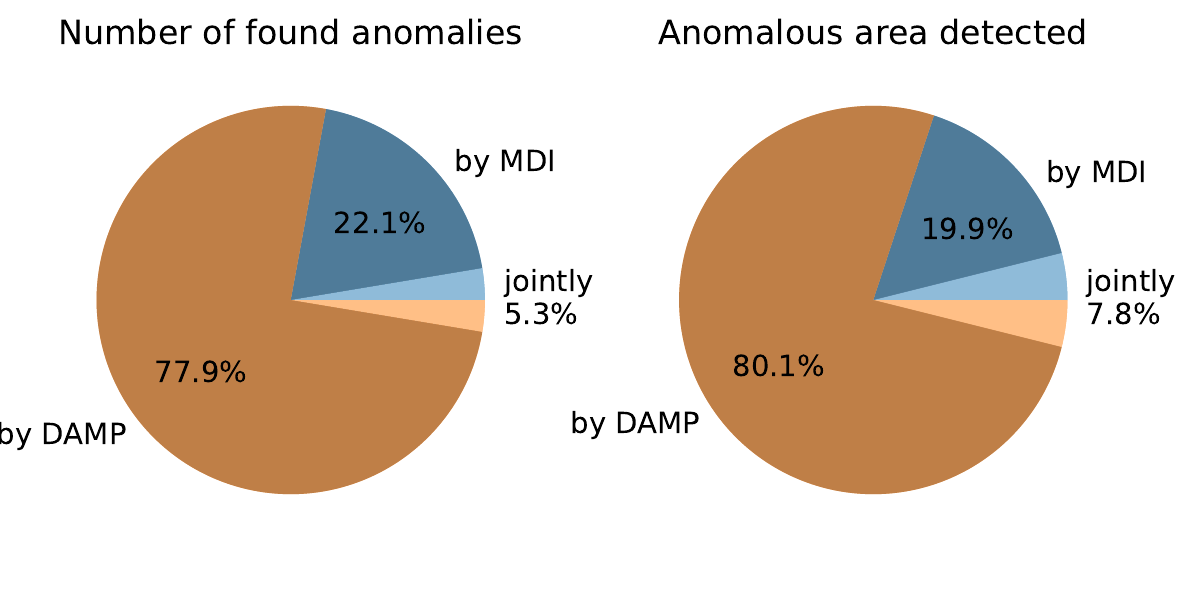}
        \caption{}
        \label{fig:complementary_univariate}
    \end{subfigure}
    \hfill
    \begin{subfigure}[b]{0.45\textwidth}
        \centering
        \includegraphics[width=\textwidth]{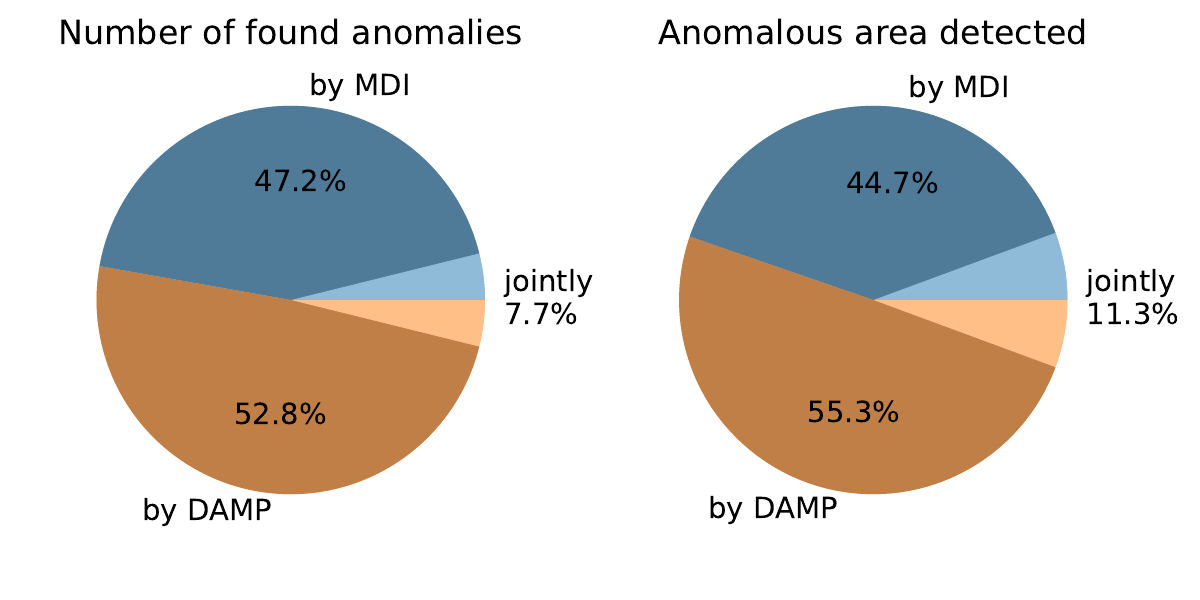}
        \caption{}
        \label{fig:complementary_multivariate}
    \end{subfigure}
    \caption{Amount of (a) univariate and (b) multivariate anomalies found by MDI and DAMP. The highlighted area marks anomalies found by both algorithms.}
    \label{fig:results:rq1}
\end{figure}

In \cite{Rewicki2023} we found, that density estimation- and discord discovery based methods , specifically MDI and MERLIN, yield complementary results in anomaly detection.
To validate this claim, we analyzed the anomalies in EDEN ISS telemetry data, found by MDI and DAMP. 
Our comparison focused on the number and duration of detected anomalies. 
The results, depicted in Fig. 1, show that DAMP accounts for the majority ($77.9\%$) of the detections in the univariate case, while $22.1\%$ of the detected subsequences are found by MDI. Only $5.3\%$ of the sequences detected as anomalous are found by MDI and DAMP simultaneously.
Considering the length of the detected subsequences (i.e. the anomalous area), only $7.8\%$ of the subsequences detected as being anomalous are detected jointly. 
In the multivariate scenario, DAMP identifies 52.8\% and MDI 47.2\% of the subsequences, with 7.7\% being detected by both. 
Even though we can not assume all detections being true positives, these findings confirm the complementary nature of MDI and DAMP, with DAMP finding slightly longer anomalies compared to MDI. Hence, using both methods allows us finding a larger variety of anomalous behavior instead of using just one.

\subsection{RQ2: Which features yield the best results?}
\label{results:rq2}
\begin{figure}[t]
    \centering
    \begin{subfigure}[b]{0.55\textwidth}
        \centering
        \includegraphics[width=\textwidth]{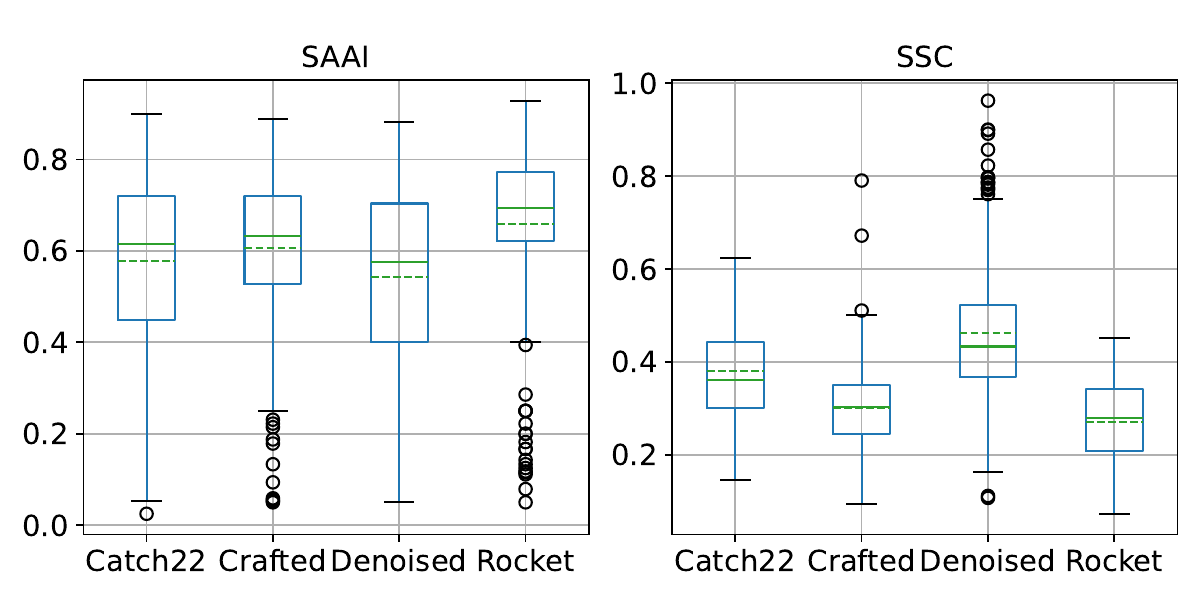}
        \caption{}
        \label{fig:feature_comparison_kmeans_uv}
    \end{subfigure}\
    \hfill
    \begin{subfigure}[b]{0.275\textwidth}
        \centering
        \includegraphics[width=\textwidth]{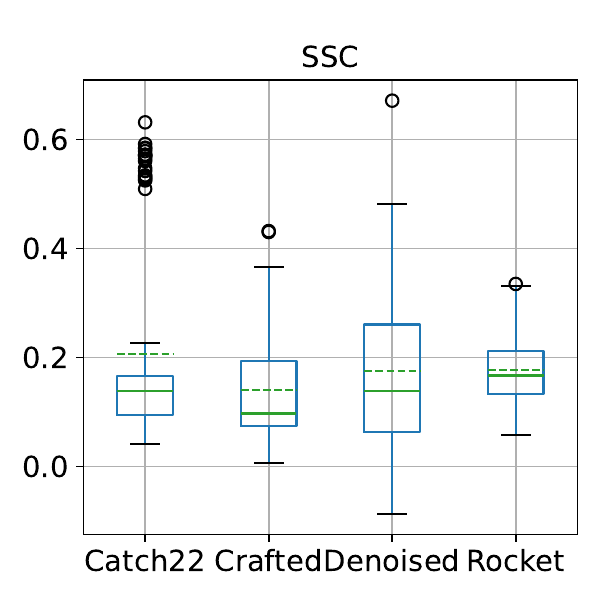}
        \caption{}
        \label{fig:feature_comparison_kmeans_mv}
    \end{subfigure}
    \hfill
    \begin{subfigure}[b]{0.55\textwidth}
        \centering
        \includegraphics[width=\textwidth]{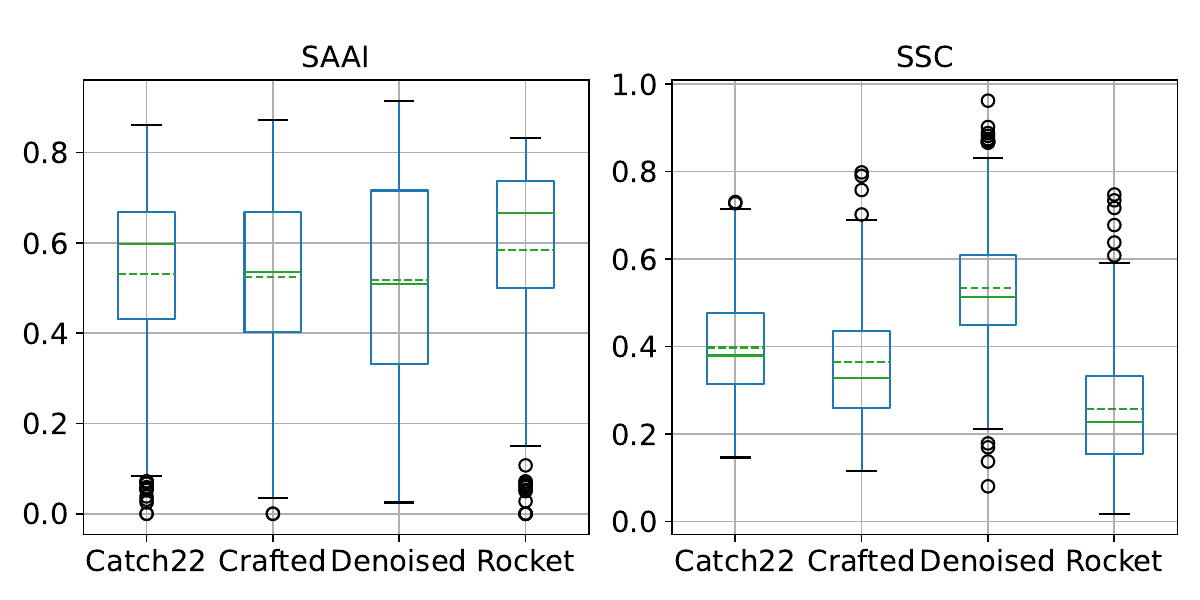}
        \caption{}
        \label{fig:feature_comparison_agglomerative_uv}
    \end{subfigure}
    \hfill
    \begin{subfigure}[b]{0.275\textwidth}
        \centering
        \includegraphics[width=\textwidth]{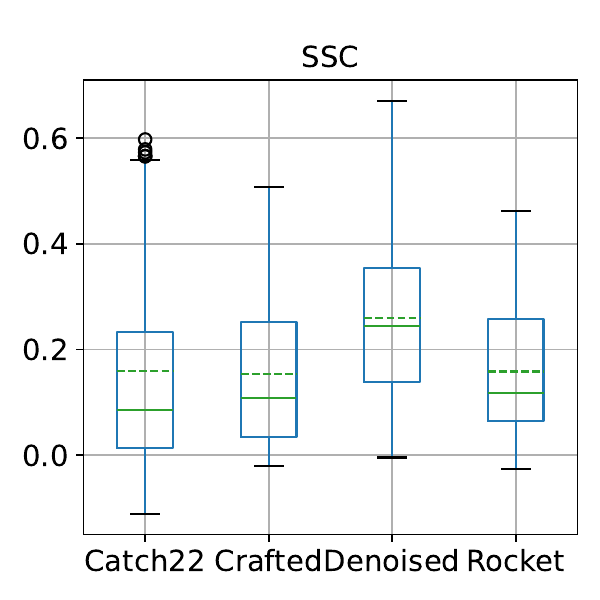}
        \caption{}
        \label{fig:feature_comparison_agglomerative_mv}
    \end{subfigure}
    \caption{(a), (c) $SAAI$ and $SSC$ aggregated over all sensor types for the four different feature extraction methods for (a) K-Means and (c) HAC for univariate anomalies. (b), (d): $SSC$ over all subsystems in the multivariate case.}
    \label{fig:results:rq2}
\end{figure}

To determine which features yield better results, we clustered anomalies for each sensor type (univariate anomalies) or subsystem (multivariate anomalies) with varying cluster numbers $K=2$ to $K=20$. We then aggregated the sensor-type or subsystem specific outcomes over the number of clusters $K$. 
We assessed the quality based on the temporal alignment of clusters using $SAAI$ (Equation \ref{eq:saai}) for univariate anomalies, and cluster separation and cohesion using $SSC$ (Equation \ref{eq:ssc}) for uni- and multivariate anomalies. 
The results are illustrated as box plots in Figure \ref{fig:results:rq2}.

\paragraph{Univariate anomalies}
Analyzing the $SAAI$ distribution for the four feature sets with K-Means clustering (Figure \ref{fig:feature_comparison_kmeans_uv}), \texttt{Rocket} features exhibit the highest median $SAAI$ of $0.69$, followed by \texttt{Crafted} features with $0.63$ and \texttt{Catch22} with $0.61$. 
\texttt{Denoised} subsequences yield a median $SAAI$ of $0.58$. 
With HAC, \texttt{Rocket} features show the highest median $SAAI$ of $0.67$, followed by \texttt{Catch22} with $0.6$, \texttt{Crafted} with $0.54$ and \texttt{Denoised} features with a median $SAAI$ of $0.51$. 
\texttt{Denoised} features display the highest variability in $SAAI$ results, while \texttt{Rocket} features demonstrate the lowest variability for both K-Means clustering and HAC.
Regarding cluster separation and cohesion (right plots in Figure \ref{fig:results:rq2}), \texttt{Denoised} features return the highest median $SSC$ of $0.43$ for K-Means and $0.51$ for HAC, while \texttt{Rocket} features yield the lowest median $SSC$ of $0.28$ for K-Means and $0.23$ for HAC.
Although the discrepancy between $SAAI$ and $SSC$ results might initially seem contradictory, it can be better understood when considering different cluster imbalances, as it will be discussed in Section \ref{results:rq3}. 
For \texttt{Denoised}, \texttt{Crafted}, and \texttt{Catch22} features, higher Silhouette scores are observed for HAC compared to K-Means, indicating higher cluster imbalances. 
Since $SSC$ is the mean of Silhouette Coefficients and does not consider cluster sizes, higher $SSC$ values may result from many small but dense clusters and few large and dispersed ones, compared to a more even distribution across clusters.

\paragraph{Multivariate Anomalies}
Since $SAAI$ is not defined in the multivariate case, we evaluate $SSC$ results, aggregated across all subsystems, which are presented in Figure \ref{fig:feature_comparison_kmeans_mv} and \ref{fig:feature_comparison_agglomerative_mv}. 
For K-Means clustering, \texttt{Rocket} features exhibit the highest median $SSC$ value of $0.17$, whereas for HAC, \texttt{Denoised} features demonstrate the highest $SSC$, with a value of $0.24$.

Based on these results we favor \texttt{Rocket} features as this feature set shows superior in terms of $SAAI$ when clustering univariate anomalies and the highest median $SSC$ in the multivariate case. For HAC the results are more inconclusive.

\subsection{RQ3: Which algorithm yields better results?}
\label{results:rq3}
\begin{figure}[t]
    \centering
    \begin{subfigure}[b]{0.40\textwidth}
        \centering
        \includegraphics[width=\textwidth]{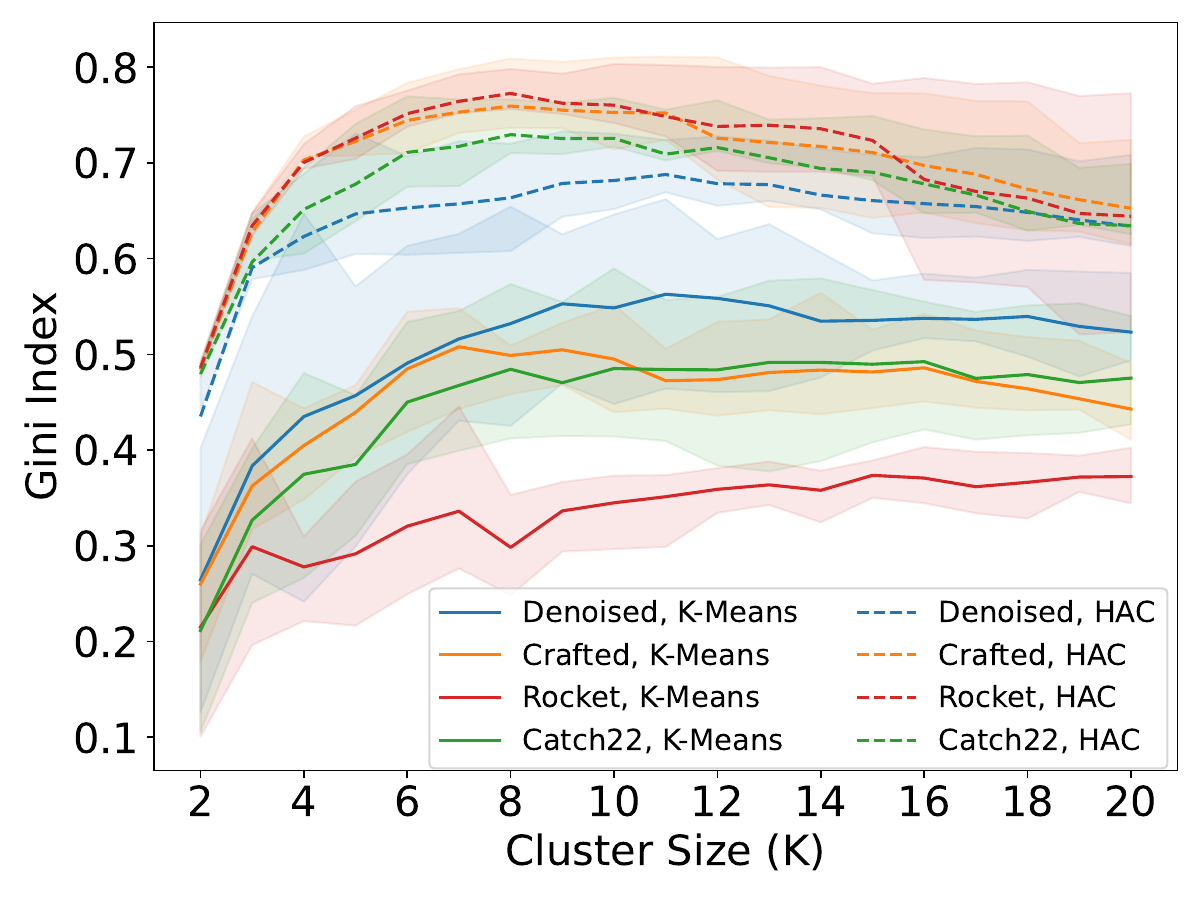}
        \caption{}
        \label{fig:sizes_gini_uv}
    \end{subfigure}
    \begin{subfigure}[b]{0.40\textwidth}
        \centering
        \includegraphics[width=\textwidth]{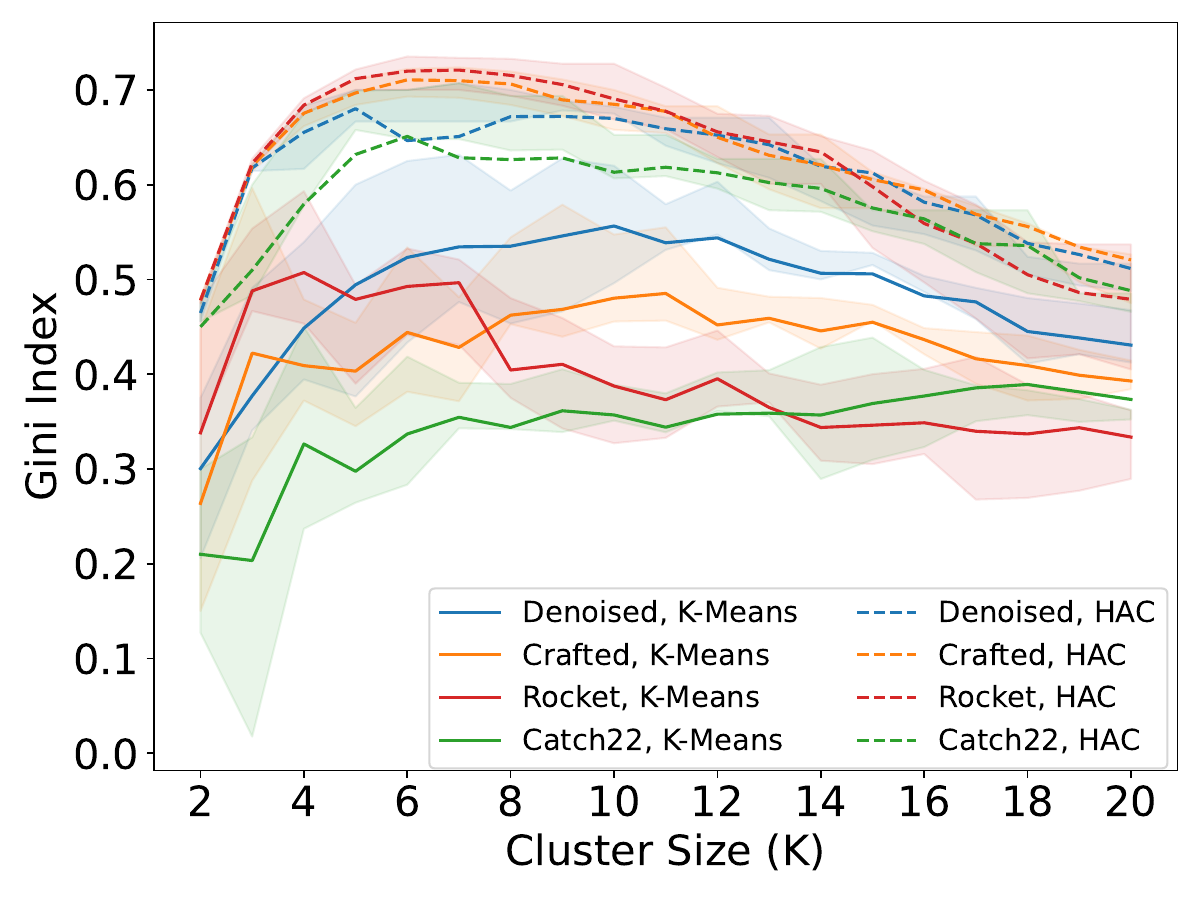}
        \caption{}
        \label{fig:sizes_gini_mv}
    \end{subfigure}
    \caption{Gini-Index of cluster sizes for K-Means Clustering and HAC in (a) the univariate and (b) the multivariate case. Lines represent mean values, while the opaque bands span the second and third quartiles.}
    \label{fig:results:rq3}
\end{figure}

To determine which algorithm yields superior results, we evaluate the imbalance of cluster sizes generated by K-Means and HAC. 
Although we cannot assume a uniform distribution of anomaly types within the EDEN ISS dataset, we prefer a more even distribution across clusters to support the isolation of varied anomalous behavior. 
Similar to Section \ref{results:rq2}, we cluster anomalies for each sensor type with increasing cluster numbers $K$ (ranging from $K=2$ to $K=20$) and aggregate the sensor-type-specific results. 
Figure \ref{fig:results:rq3} depicts the Gini-Index for increasing $K$ by feature set and clustering algorithm. 
It is evident from the results, that K-Means produces more evenly distributed clusters compared to HAC for all four feature sets.

\paragraph{Univariate Anomalies}
K-Means shows moderate imbalance, with average Gini Indices up to $0.51$ for \texttt{Denoised} features, while HAC yields higher Gini Indices, averaging up to $0.77$ for \texttt{Rocket}. 
In K-Means clustering, \texttt{Rocket} features generate the most balanced clusters, whereas \texttt{Denoised} features exhibit the highest imbalance. 
Conversely, for HAC, this observation is inverted. 
The trajectories of the Gini-Index curves are largely similar within each algorithm. 
For K-Means clustering, Gini Indices increase until $K=11$, while for HAC, they peak around $K=8$ before slowly decreasing. 
This suggests that in HAC, large clusters are not split but gradually dissolve as $K$ increases. 
\texttt{Rocket} with K-Means clustering follows a slightly different trajectory, slowly increasing until $K=20$.

\paragraph{Multivariate Anomalies}
In comparison to HAC, K-Means clustering shows more balanced cluster sizes, consistent with the findings in the univariate case. 
\texttt{De}-\texttt{noised} features have the highest average Gini Index of $0.56$ for K-Means, while for HAC, \texttt{Rocket} features exhibit the highest mean value of $0.72$. 
The trajectories of the Gini Index curves are largely similar for each algorithm, though for HAC, there is a steeper decline from $K=8$, especially noticeable when compared to the univariate case. 

The observation of K-Means producing more balanced cluster sizes, both in univariate and multivariate scenarios, prompts us to focus on addressing the remaining two research questions w.r.t. K-Means.

\subsection{RQ4: What anomaly types can be isolated?}
\label{results:rq4}
To analyze, which anomaly types can be isolated from the clustering results, we analyze the consensus between the $SAAI$-optimal clustering solutions for the four different feature sets. 
Given the clustering result $C_A = \lbrace{C_{1_A}, C_{2_A}, \dots, C_{K_A}\rbrace}$ for a feature set $A$ and $C_B = \lbrace{C_{1_B}, C_{2_B}, \dots, C_{K_B}\rbrace}$ a feature set $B$, we calculate the matrix of pairwise intersection over union
$\mathbf{M}_{AB}(i,j) = iou(C_{i_A}, C_{j_B})$
. 
The values in $\mathbf{M}_{AB}$ are normalized to the interval $[0, 1]$.
We consider cluster $i$ in feature set $A$ to isolate the same anomaly type as cluster $j$ in feature set $B$, if $\mathbf{M}_{AB}(i,j) \geq 0.5$, i.e. if both clusters share at least $50\%$ of their samples.

\begin{figure}[t]
    \centering
    \begin{subfigure}[b]{.524\textwidth}
        \centering
        \includegraphics[width=\textwidth]{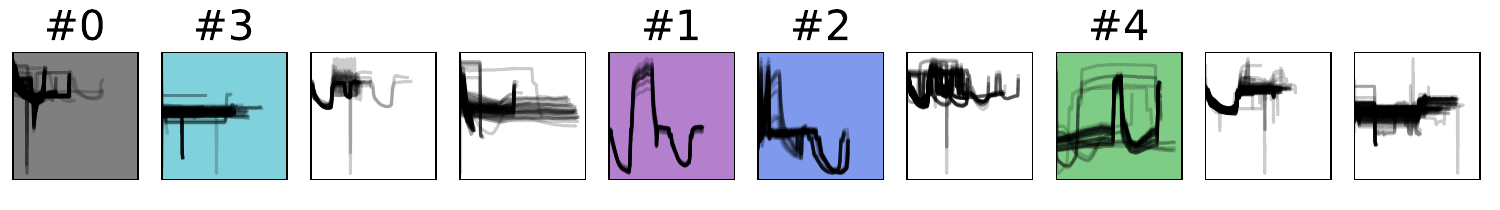}
        \label{fig:ics_denoised_10}
    \end{subfigure}
    \hfill
    \begin{subfigure}[b]{.68\textwidth}
        \centering
        \vspace{-13px}
        \includegraphics[width=\textwidth]{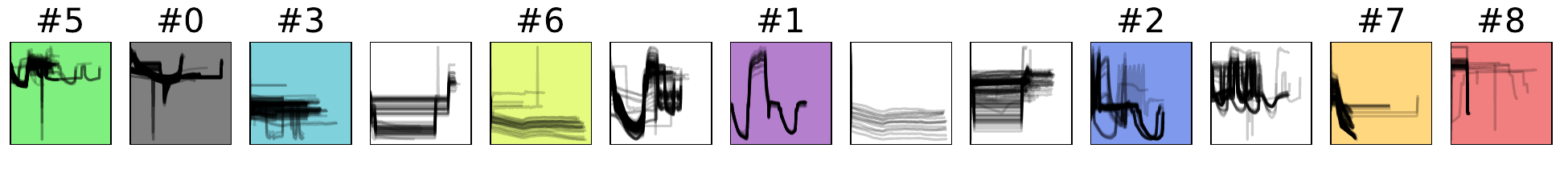}
        \label{fig:ics_crafted_13}
    \end{subfigure}
    \hfill
    \begin{subfigure}[b]{\textwidth}
        \centering
        \vspace{-13px}
        \includegraphics[width=\textwidth]{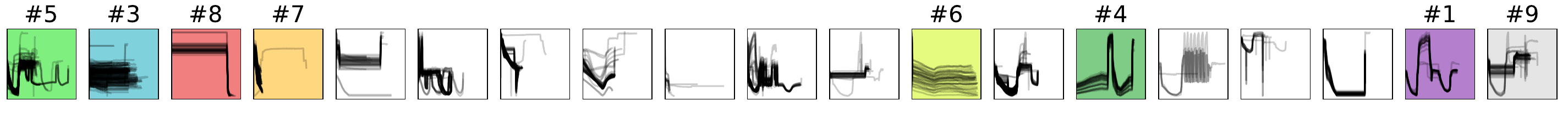}
        \label{fig:ics_rocket_19}
    \end{subfigure}
    \hfill
    \begin{subfigure}[b]{.42\textwidth}
        \centering
        \vspace{-13px}
        \includegraphics[width=\textwidth]{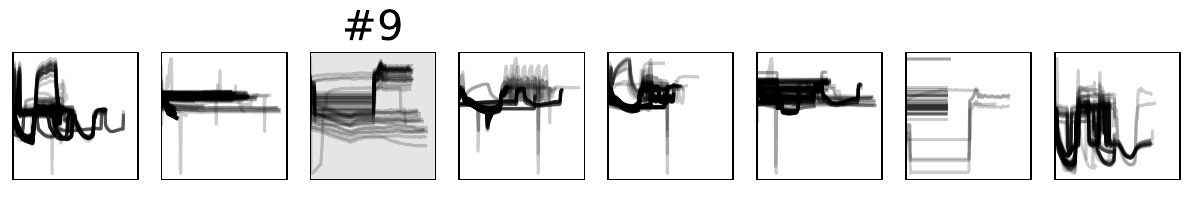}
        \label{fig:ics_catch22_8}
    \end{subfigure}
    \caption{Anomaly types found in ICS temperatures with K-Means clustering for $SAAI$-optimal $K$ per feature set. Anomaly types consistent across different feature sets have the same color. T/B: \texttt{Denoised} \texttt{Crafted}, \texttt{Rocket}, \texttt{Catch22}}
    \label{fig:results:rq4_ics_kopt}
\end{figure}

\subsubsection{Case Study 1: Univariate Anomalies in ICS}
Figure \ref{fig:results:rq4_ics_kopt} depicts the anomaly clusters obtained from K-Means clustering with $SAAI$-optimal number of clusters $K$\footnote{High-Resolution versions of the images can be found in the supplemental materials~\ref{appendix:anomaly_types}}. The $SAAI$ results for each feature set and $2 \leq K \leq 20$ are shown in the supplemental Figure \ref{fig:results:rq4_ics_saai}.
The ICS measurements consist of 38 time series, representing temperature readings below the LED lamps for each growth tray, with 1303 identified anomalies for this sensor type.
Optimal results for \texttt{Catch22} and \texttt{Denoised} features were observed at $K=8$ and $K=10$, while for \texttt{Crafted} and \texttt{Rocket} features, the highest $SAAI$ values were obtained at $K=13$ and $K=19$. 
Anomaly type candidates derived from the consensus criterion are highlighted with colored background in Figure \ref{fig:results:rq4_ics_kopt}.
Table \ref{tab:rq4_uv} in the supplemental materials~\ref{appendix:anomaly_types} provides shape descriptions for 10 isolated anomaly type candidates identified in ICS temperature readings.
The "peak (long)" (\#1) anomalies are isolated by \texttt{Denoised}, \texttt{Crafted}, and \texttt{Rocket} features. The remaining anomaly type candidates were isolated by two feature sets.
\texttt{Rocket} and \texttt{Crafted} features yield the most anomaly type candidates, reflecting their highest $SAAI$ for the highest number of clusters. For candidate \#7, we found no indication of anomalous behavior, so we suspect that this cluster contains false positives results.
To investigate whether a larger number of clusters alone aids in analysis, we analyzed the isolated anomaly types at $K=19$ for all feature sets.
Annotated sequence cluster plots are presented in the supplemental Figure \ref{fig:results:rq4_ics_k19}. 
\texttt{Rocket} features isolate the most, i.e. $8$, distinct anomaly type candidates, followed by \texttt{Crafted}  and \texttt{Catch22} with $6$, and \texttt{Denoised} with $4$ distinct anomaly type candidates, indicating superior performance for \texttt{Rocket} in forming interpretable anomaly clusters despite the increase in $K$. \texttt{Crafted} features however yielded more distinct clusters for the $SAAI$-optimal value $K=13$, underlining the efficacy of that measure.

\subsubsection{Case Study 2: Multivariate Anomalies in AMS-SES}
\label{results:rq4:mv}

\begin{figure}[t]
    \centering
    \begin{subfigure}[b]{.526\textwidth}
        \centering
        \includegraphics[width=\textwidth]{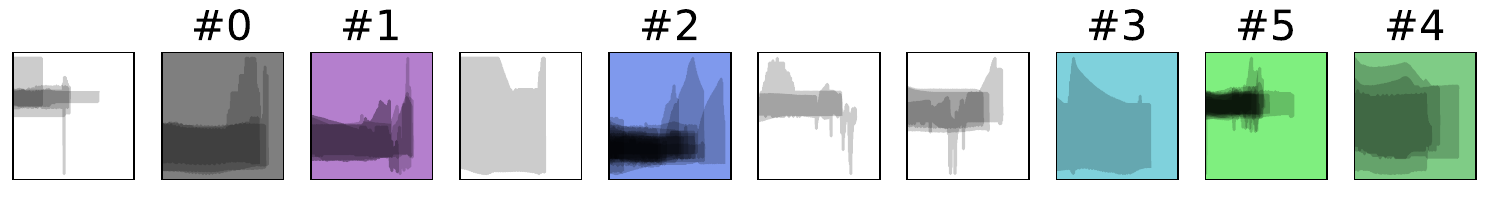}
        \label{fig:ams-ses_denoised_10}
    \end{subfigure}\\
    \vspace{-13px}
    \begin{subfigure}[b]{.526\textwidth}
        \centering
        \includegraphics[width=\textwidth]{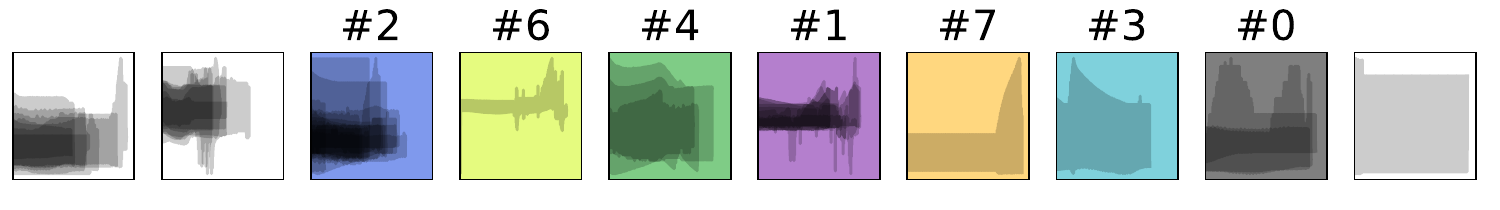}
        \label{fig:ams-ses_crafted_10}
    \end{subfigure}\\
    \vspace{-13px}
    \begin{subfigure}[b]{.526\textwidth}
        \centering
        \includegraphics[width=\textwidth]{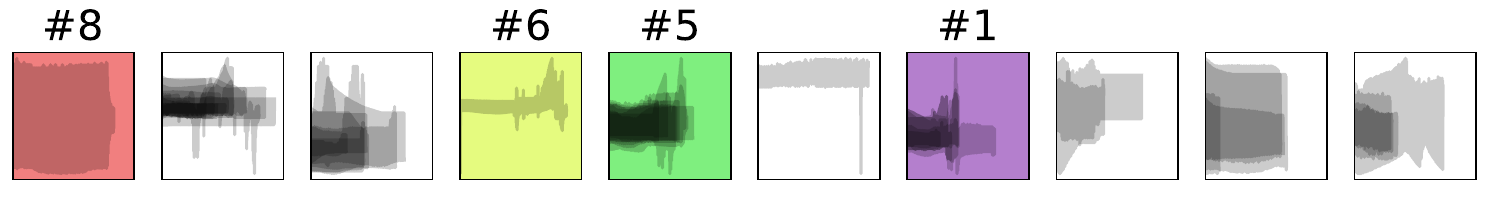}
        \label{fig:ams-ses_rocket_10}
    \end{subfigure}\\
    \vspace{-13px}
    \begin{subfigure}[b]{.526\textwidth}
        \centering
        \includegraphics[width=\textwidth]{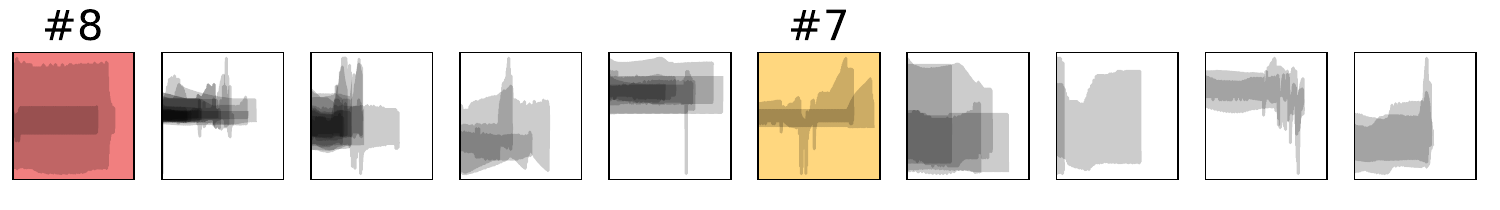}
        \label{fig:ams-ses_catch22_10}
    \end{subfigure}
    \caption{Anomaly types found in the multivariate AMS-SES measurements with K-Means clustering for $K=10$. T/B: \texttt{Denoised}, \texttt{Crafted}, \texttt{Rocket}, \texttt{Catch22}}
    \label{fig:results:rq4_ams-ses_k10}
\end{figure}

In the multivariate case, we examine anomalies within each subsystem, here focusing on the AMS-SES subsystem. 
As $SAAI$ computation for multivariate anomalies is not feasible, we fix $K=10$ and use the consensus criterion to delineate anomaly type candidates from clustering results. 
Figure \ref{fig:results:rq4_ams-ses_k10} displays the outcomes, with \texttt{Crafted} features discerning the majority (i.e. $7$) of anomaly type candidates, followed by \texttt{Denoised} features with $6$. 
\texttt{Rocket} features isolate $4$ candidates and \texttt{Catch22} only $2$. 
The "$CO^2$ Peak" anomaly (\#1) exhibits the highest consensus, detected by \texttt{Denoised}, \texttt{Crafted}, and \texttt{Rocket} features. 
All other anomaly type candidates show a consensus of $2/4$. 
Descriptions for isolated anomaly type candidates are provided in Table \ref{tab:rq4_mv} in the supplemental materials. 
While semantic interpretation remains elusive, we characterize them based on their shape. 
For instance, anomaly types (\#5, \#8) in Figure \ref{fig:results:rq4_ams-ses_k10} denote various manifestations of "steep slope" anomalies. 
Yet, descriptions for candidates (\#0, \#2, \#3) were challenging, hinting at potential false positive anomaly detection results. 
In summary, interpreting multivariate anomalies is more challenging due to diverse sensor readings in each subsystem's multivariate time series.

\subsection{RQ5: Can we identify recurring abnormal behavior?}
\label{results:rq5}

\begin{figure}[t]
    \centering
    \begin{subfigure}[b]{0.24\textwidth}
        \centering
        \includegraphics[width=\textwidth]{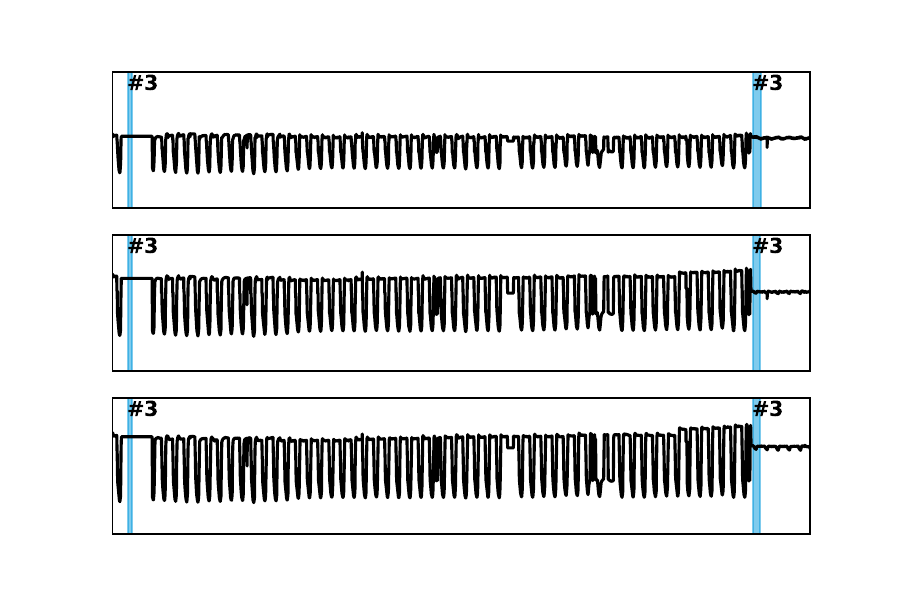}
        \caption{}
        \label{fig:rq5:ics:3}
    \end{subfigure}\
    \hfill
    \begin{subfigure}[b]{0.24\textwidth}
        \centering
        \includegraphics[width=\textwidth]{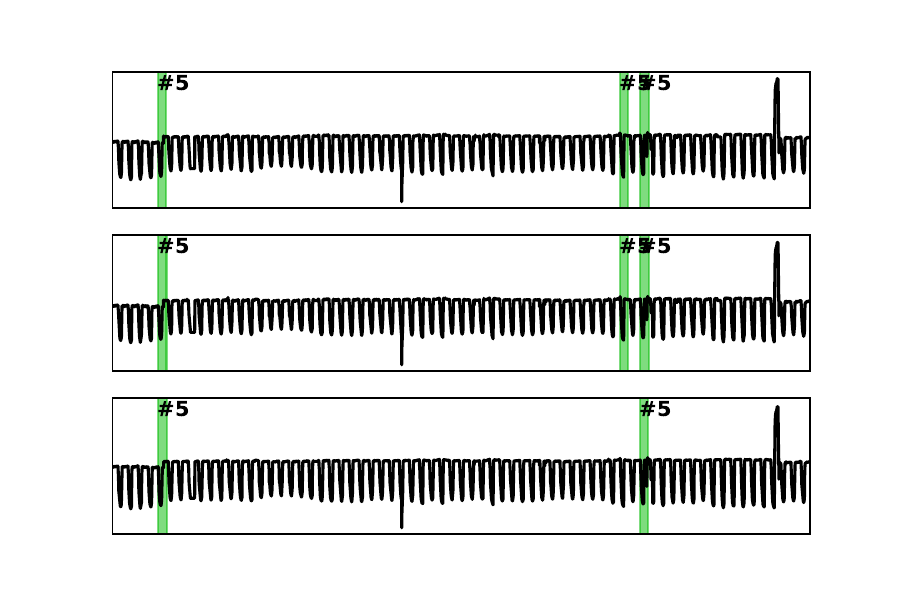}
        \caption{}
        \label{fig:rq5:ics:5}
    \end{subfigure}
    \hfill
    \begin{subfigure}[b]{0.24\textwidth}
        \centering
        \includegraphics[width=\textwidth]{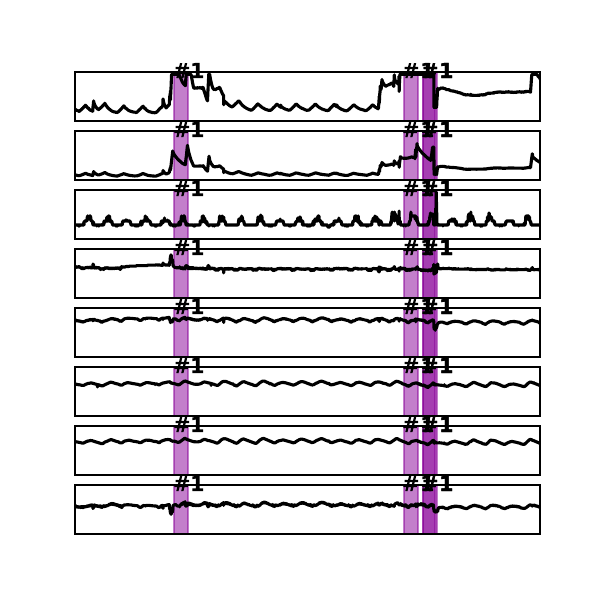}
        \caption{}
        \label{fig:rq5:ams-ses:1}
    \end{subfigure}
    \begin{subfigure}[b]{0.24\textwidth}
        \centering
        \includegraphics[width=\textwidth]{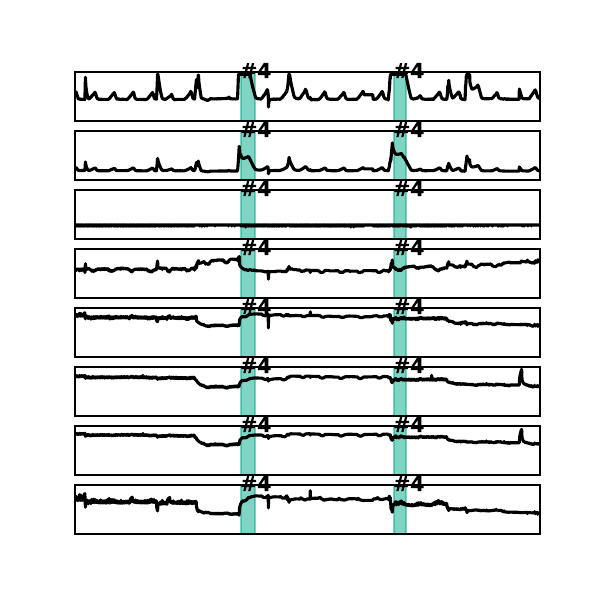}
        \caption{}
        \label{fig:rq5:ams-ses:4}
    \end{subfigure}
    \caption{Recurring  anomalies in the univariate ICS readings: (a) Near-flat or Flat Signal (\#3), (b) Anomalous Day Phase (\#5) and in the AMS-SES measurements: (c) $CO^2$ Peak (\#1), (d) Temp. and $CO^2$ Peak with RH Drop (\#4).}
    \label{fig:results:rq5:ics}
\end{figure}

To identify recurring abnormal behavior, we focus on anomaly types with multiple instances. 
In Figure \ref{fig:results:rq5:ics}, we illustrate examples of these type candidates for the univariate (\ref{fig:rq5:ics:3}, \ref{fig:rq5:ics:5}) and multivariate case (\ref{fig:rq5:ams-ses:1}, \ref{fig:rq5:ams-ses:4}). 
While the univariate "near flat or flat signal" anomaly type (\#3) shows nearly identical behavior across both instances, the "anomalous day phase" type (\#5) exhibits greater diversity. 
In the initial instance, the warm-up phase involves a step increase followed by a rise in daytime temperature, while in the third instance of the "anomalous day phase" anomaly, a daytime drop occurs after achieving the target temperature.

Using the same methodology in the multivariate case for AMS-SES readings as described in Section \ref{results:rq4:mv}, we present two instances of recurring anomalous behavior, namely anomaly types "$CO^2$ peak" (\#1) and "temperature and $CO^2$ peak with relative humidity drop" (\#4), in Figure \ref{fig:rq5:ams-ses:1} and \ref{fig:rq5:ams-ses:4}. 
In both cases, the anomaly is observable and shows similar behavior across the instances, but we lack evidence to assert consistent underlying causes across instances. 
Labeling these instances as the same anomaly type requires a more in-depth analysis of the anomalies and their root causes, a task we leave for future research.


\section{Conclusions and Outlook}
\label{sec:conclusion}
In this study, we analyzed anomalies in telemetry data from the BLSS prototype, EDEN ISS, during the mission year 2020. 
Using anomaly detection methods MDI and DAMP, we extracted four feature sets from identified anomalous subsequences. Employing K-Means clustering and HAC, we aimed to isolate various anomaly types. 
Our findings showed K-Means produced more uniform cluster sizes compared to HAC, aligning with our goal of discerning diverse anomalous behavior forms. 
We found \texttt{Rocket} and \texttt{Crafted} features outperformed \texttt{Denoised} subsequences and \texttt{Catch22} features in detecting univariate anomalies. 
However, assessing multivariate anomalies quality solely using $SSC$ proofed challenging. Despite these challenges, our analysis identified various anomaly types, including peaks, anomalous day/night patterns, drops, and delayed events, through consensus among different feature sets. These insights are crucial for refining our risk mitigation system for future BLSS iterations. We identified instances of potentially recurring anomalous behavior in both uni- and multivariate contexts, warranting further investigation. Additionally, we will further explore Catch22 features, promising informative insights into our problem domain.

\subsubsection*{Acknowledgments}
The authors would like to thank Vincent Vrakking and Daniel Schubert for providing the raw data and Tom Norman for compiling the dataset used in this work. We also thank the anonymous reviewers for their valuable suggestions.

\subsubsection*{Disclosure of Interests}
The authors have no competing interests to declare that are relevant to the content of this article.

\clearpage
\begin{center}
\textbf{\large Supplemental Materials: Unraveling Anomalies in Time: Unsupervised Discovery and Isolation of Anomalous Behavior in Bio-regenerative Life Support System Telemetry}
\end{center}
\setcounter{equation}{0}
\setcounter{figure}{0}
\setcounter{table}{0}
\setcounter{section}{0}
\makeatletter
\renewcommand{\theequation}{S\arabic{equation}}
\renewcommand{\thefigure}{S\arabic{figure}}
\renewcommand{\bibnumfmt}[1]{[S#1]}
\renewcommand{\citenumfont}[1]{S#1}

\appendix


\section[\appendixname~\thesection]{Examples and additional information for the Synchronized Anomaly Agreement Index (SAAI)}
\label{appendix:saai_intuition}
To support the interpretation of the SAAI, we provide examples to illustrate its computation and the relation of the main and regularizing terms.

\begin{figure}[ht]
    \centering
    \includegraphics[width=.8\textwidth]{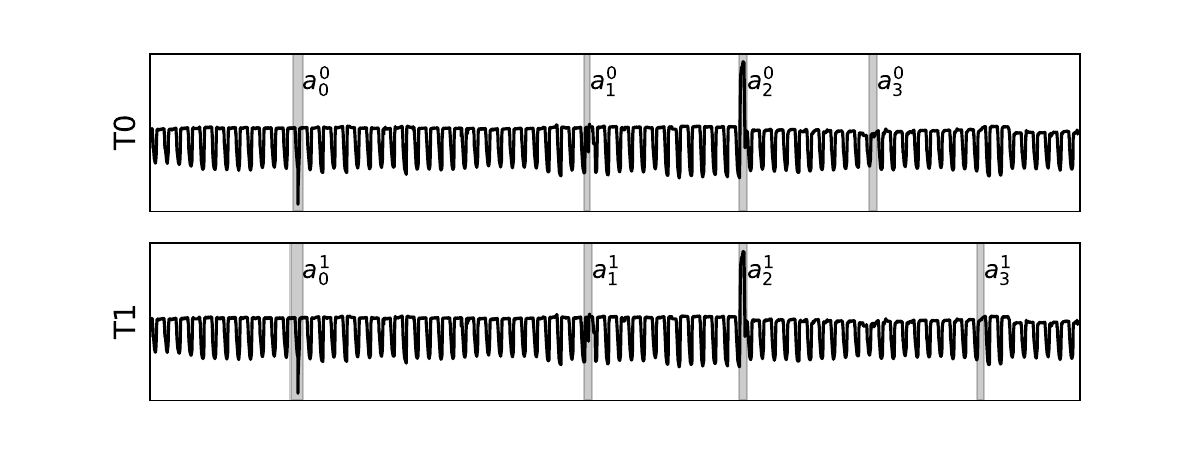}
    \caption{Time Series $\mathcal{T}$ with Anomalies $A = \lbrace{ a^i_j | i \in \lbrace{ 0,1 \rbrace}, j \in \lbrace{ 0, 1 ,2 \rbrace} \rbrace}$}
    \label{fig:appendix:saai:1}
\end{figure}

Given the regular multivariate time series
$$
\mathcal{T} = \lbrace{(t_n, \mathbf{x_n}) | t_n \leq t_{n+1}, 0 \leq n \leq N-1, t_{n+1} - t_n = c \rbrace}$$
with $p_i \in \mathbb{R}^2$ shown in Figure \ref{fig:appendix:saai:1}. Highlighted are the univariate anomalies $A = \lbrace{ a^i_j | i \in \lbrace{ 0,1 \rbrace}, j \in \lbrace{ 0, 1, 2, 3 \rbrace} \rbrace}$ that have been found in the Anomaly Detection step.

The SAAI is calculated over the set of temporally aligned anomalies:
$$
    A_S = \lbrace{(a^0_0, a^1_0), (a^0_1, a^1_1), (a^0_3, a^1_3) \rbrace}, |A_S| = 3 \text{   .}
$$
The anomalies $a^0_4$ and $a^1_4$ are not temporally aligned and hence $((a^0_4, a^1_4)) \notin A_S$.
\\\\

\begin{figure}[ht]
    \centering
    \includegraphics[width=.8\textwidth]{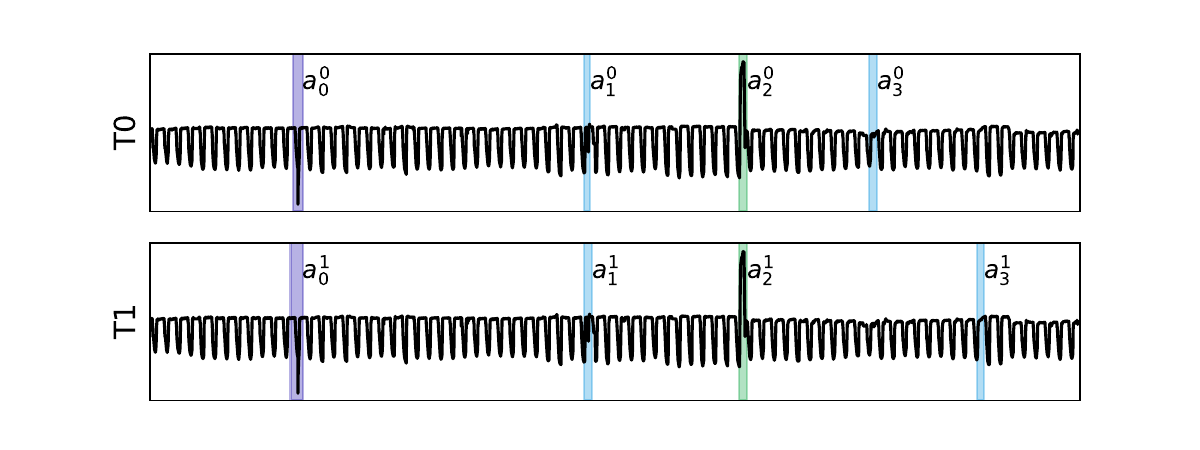}
    \caption{Temporally aligned anomalies are assigned to the same cluster, different anomaly types form different clusters, and no clusters with a single element.  $SAAI = 0.8\Bar{3}$}
    \label{fig:appendix:saai:2}
\end{figure}

We will now present different possible clustering solutions and how the SAAI is calculated in the respective situation, starting with the best case in this situation, shown in Figure \ref{fig:appendix:saai:2}.

The SAAI is calculated as: 
\begin{align*}
k &= 3, n_\mathds{1} = 0  \\ 
A^*_S &= \lbrace{(a^0_0, a^1_0), (a^0_1, a^1_1), (a^0_3, a^1_3) \rbrace},  |A^*_S| = 3 \\
SAAI &:= \lambda \frac{|A^*_S|}{|A_S|} - (1-\lambda)(\frac{1}{k} + \frac{n_\mathds{1}}{k}) + (1-\lambda) = 0.5 \cdot \frac{3}{3} - 0.5 (\frac{1}{3} + \frac{0}{3}) + 0.5 = 0.8\Bar{3}    
\end{align*}

\begin{figure}[ht]
    \centering
    \includegraphics[width=.8\textwidth]{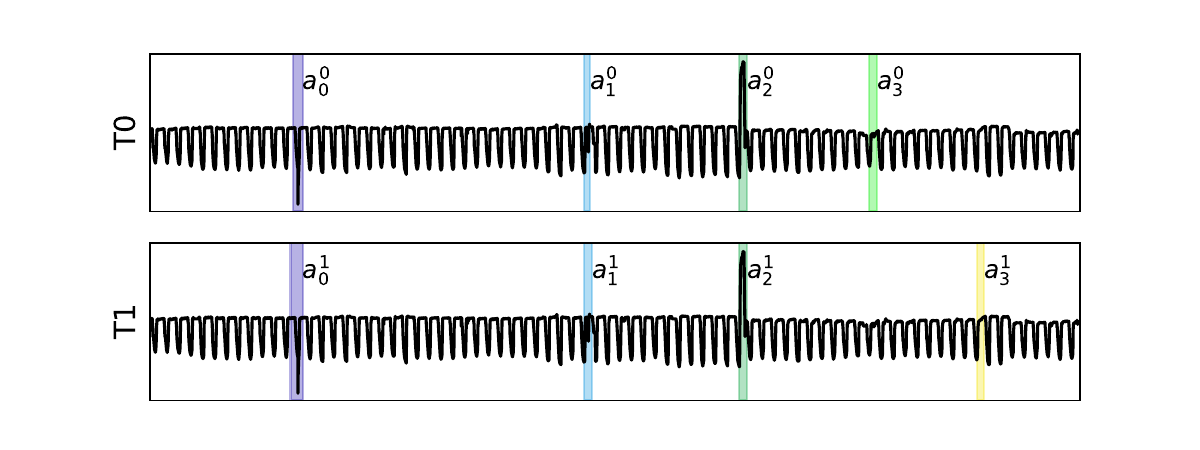}
    \caption{Temporally aligned anomalies are assigned to the same cluster, different anomaly types form different clusters but 2 clusters with a single element.  $SAAI = 0.7$}
    \label{fig:appendix:saai:3}
\end{figure}

The second example shown in Figure \ref{fig:appendix:saai:3} is similar to the situation in Figure \ref{fig:appendix:saai:2}, but the unaligned anomalies $a^0_3$ and $a^1_3$ are assigned to clusters that contain no other element. This increases the number of clusters on one hand but increases the number of clusters with a single element on the other hand. Both are captured by the regularization term as visible in Equation \ref{eq:saai:3}.

\begin{align}
k &= 5,   n_\mathds{1} = 2 \notag \\
A^*_S &= \lbrace{(a^0_0, a^1_0), (a^0_1, a^1_1), (a^0_3, a^1_3) \rbrace}, |A^*_S| = 3 \notag \\
SAAI &:= \lambda \frac{|A^*_S|}{|A_S|} - (1-\lambda)(\frac{1}{k} + \frac{n_\mathds{1}}{k}) + (1-\lambda) = 0.5 \cdot \frac{3}{3} - 0.5 (\frac{1}{5} + \frac{2}{5}) + 0.5 = 0.7
\label{eq:saai:3}
\end{align}

\begin{figure}[ht]
    \centering
    \includegraphics[width=.8\textwidth]{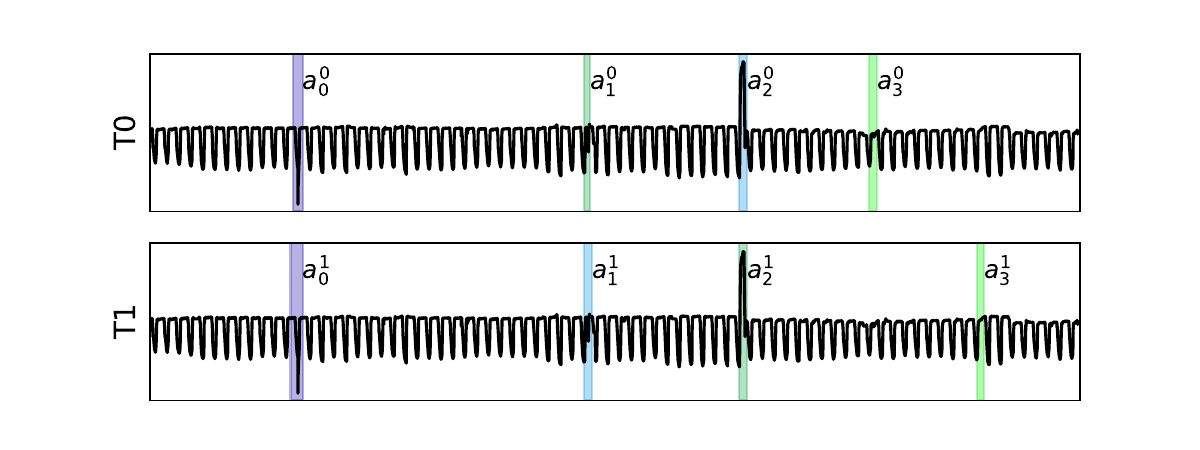}
    \caption{Only one pair of temporally aligned anomalies is assigned to the same cluster but not singular clusters $SAAI = 0.541\bar{6}$}
    \label{fig:appendix:saai:4}
\end{figure}

In the solution depicted in Figure \ref{fig:appendix:saai:4}, only one pair of temporally aligned anomalies $(a^0_0, a^1_0)$ is assigned to the same cluster, which is worse compared to the solution in Figure \ref{fig:appendix:saai:3}, but at least one anomaly type is still isolated.

\begin{align*}
k &= 4,   n_\mathds{1} = 0 \\
A^*_S &= \lbrace{(a^0_0, a^1_0) \rbrace}, |A^*_S| = 1 \\
SAAI &:= \lambda \frac{|A^*_S|}{|A_S|} - (1-\lambda)(\frac{1}{k} + \frac{n_\mathds{1}}{k}) + (1-\lambda) = 0.5 \cdot \frac{1}{3} - 0.5 (\frac{1}{4} + \frac{0}{4}) + 0.5 = 0.541\bar{6}
\end{align*}

\begin{figure}[ht]
    \centering
    \includegraphics[width=.8\textwidth]{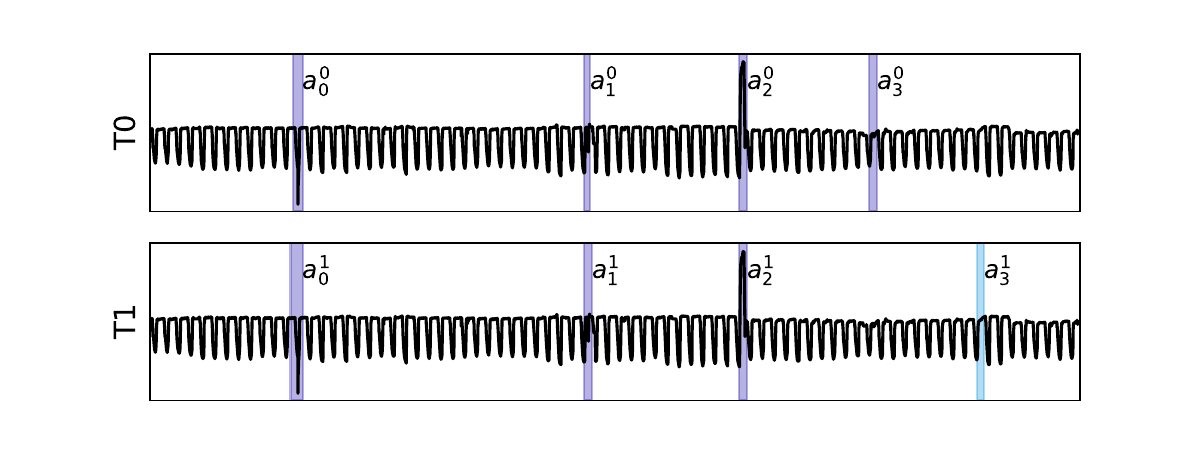}
    \caption{All Temporally aligned anomalies are assigned to the same cluster of 2 clusters. The other cluster contains only one element  $SAAI = 0.5$}
    \label{fig:appendix:saai:5}
\end{figure}

In the example shown in Figure \ref{fig:appendix:saai:5}, all temporally aligned anomalies are assigned to the same cluster. Only the unaligned anomaly $a^1_3$ is assigned to a different cluster. Although the number of singular clusters is smaller than in the Solution described in Figure \ref{fig:appendix:saai:3}, the solution in Figure \ref{fig:appendix:saai:5} is worse as the anomaly types are not distinguishable here.

\begin{align*}
k &= 2,   n_\mathds{1} = 1 \\
A^*_S &= \lbrace{(a^0_0, a^1_0), (a^0_1, a^1_1), (a^0_3, a^1_3) \rbrace}, |A^*_S| = 3 \\
SAAI &:= \lambda \frac{|A^*_S|}{|A_S|} - (1-\lambda)(\frac{1}{k} + \frac{n_\mathds{1}}{k}) + (1-\lambda) = 0.5 \cdot \frac{3}{3} - 0.5 (\frac{1}{2} + \frac{1}{2}) + 0.5 = 0.5
\end{align*}

\begin{figure}[ht]
    \centering
    \includegraphics[width=.8\textwidth]{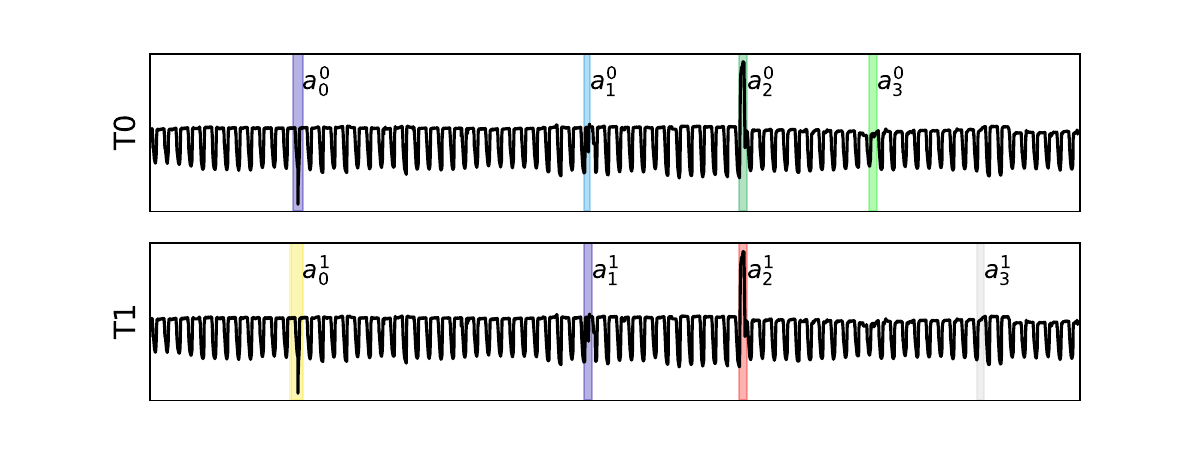}
    \caption{Worst case: No pair of temporally aligned anomalies is assigned to the same cluster and the solution contains only one cluster with more than one element.  $SAAI = 0$}
    \label{fig:appendix:saai:6}
\end{figure}

The solution illustrated in Figure \ref{fig:appendix:saai:6} shows the worst-case scenario: no pair of temporally aligned anomalies is assigned to the same cluster. Additionally, only one cluster contains more than 1 element.

\begin{align*}
k &= 7,   n_\mathds{1} = 6 \\
A^*_S &= \emptyset, |A^*_S| = 0  \\
SAAI &:= \lambda \frac{|A^*_S|}{|A_S|} - (1-\lambda)(\frac{1}{k} + \frac{n_\mathds{1}}{k}) + (1-\lambda) = 0.5 \cdot \frac{0}{3} - 0.5 (\frac{1}{7} + \frac{6}{7}) + 0.5 = 0
\end{align*}

\clearpage

\section[\appendixname~\thesection]{Hyperparameter settings for the experiments}
\label{appendix:parameters}

\begin{longtable}{llll}
\toprule
\textbf{Method} &
  \textbf{Parameter} &
  \textbf{Value} &
  \textbf{Note} 
\endhead
\midrule
\textbf{MDI} &
  $L_{min}$ &
  144 &
  \begin{tabular}[t]{@{}l@{}}Minimum Length of anomalous Intervals.\\144 corresponds to 0.5 days. Set empirically.\end{tabular}\\
 &
  $L_{max}$ &
  288 &
  \begin{tabular}[t]{@{}l@{}}Maximum Length of anomalous Intervals.\\288 corresponds to a full day. Set empirically.\end{tabular}\\
 &
  method &
  gaussian\_cov &
  Method for probability density estimation \\
 &
  mode &
  TS &
  \begin{tabular}[t]{@{}l@{}}Divergence to be used. \\ TS $\hat{=}$ unbiased KL divergence\end{tabular} \\
 &
  proposals &
  hotellings\_t &
  Interval proposal method \\
 &
  preproc &
  td &
  \begin{tabular}[t]{@{}l@{}}Preprocessing: Time-Delay Embedding \\ with automatic  determination of the \\ embedding dimension\end{tabular} \\ \midrule
\textbf{DAMP} &
  m &
  288 &
  Subsequence length. 288 corresponds to a full day. \\
 &
  t0 &
  2880 &
  \begin{tabular}[t]{@{}l@{}}Location of split point between training and \\ test data (spIndex)\end{tabular} \\
 &
  lookahead &
  0 &
  \begin{tabular}[t]{@{}l@{}}Length to "peak" ahead when pruning in the \\ forward pass\end{tabular} \\ \midrule
\textbf{Denoised} &
  windowsize &
  5 &
  Length of the moving average window. \\ \midrule
\textbf{Rocket} &
  num\_kernels &
  1000 &
  Number of random kernels \\
 &
  normalize &
  True &
   \\
 &
  pca\_components &
  10 &
  Number of principal components to use as features \\ \midrule
\textbf{Catch22} &
  th &
  \begin{tabular}[t]{@{}l@{}}0.01\\ 0.0001\end{tabular} &
  \begin{tabular}[t]{@{}l@{}}Variance threshold for selecting features \\ from catch22.  $0.01$ is used in the univariate \\ and $0.0001$ in the multivariate case.\end{tabular} \\ \midrule
\textbf{K-Means} &
  metric &
  \begin{tabular}[t]{@{}l@{}}dtw\\ euclidean\end{tabular} &
  \begin{tabular}[t]{@{}l@{}}Metric to use for K-Means clustering: "dtw" is used \\ with \texttt{Denoised} features and "euclidean" with \\ \texttt{Crafted}, \texttt{Rocket} and  \texttt{Catch22}\end{tabular} \\ \midrule
\textbf{HAC} &
  metric &
  \begin{tabular}[t]{@{}l@{}}dtw\\ euclidean\end{tabular} &
  \begin{tabular}[t]{@{}l@{}}Metric to use for HAC: "dtw" is used with \\ \texttt{Denoised} features and  "euclidean" with \\ \texttt{Crafted}, \texttt{Rocket} and  \texttt{Catch22}.\end{tabular} \\
 &
  linkage &
  centroid &
  Method for calculating the distance between clusters \\
 &
  criterion &
  maxclust &
  The criterion to use in forming flat clusters. \\ \midrule
\textbf{SAAI} &
  $\lambda$ &
  0.5&
  Weight for calculating the SAAI \\ \midrule
\textbf{Misc} &
  seed &
  42 &
  Seed to use for random number generation \\ \bottomrule
\label{tab:hyperparams}
\end{longtable}

\clearpage

\section[\appendixname~\thesection]{EDEN ISS Subsystems and their sensors}
\label{appendix:dataset_subsystems}
\begin{table}[ht]
\centering
\begin{tabular}{@{}lll@{}}
\toprule
\textbf{Subsystem} & \textbf{Sensor}                                                                                                                                               & \textbf{\# Sensors}                                           \\ \midrule
AMS                & \begin{tabular}[c]{@{}l@{}}CO2\\ Photosynthetic Active Radiation (PAR)\\ Relative Humidity (RH)\\ Temperature (T)\\ Vapor Pressure Deficit (VPD)\end{tabular} & \begin{tabular}[c]{@{}l@{}}4\\ 3\\ 3\\ 6\\ 1\end{tabular}     \\ \midrule
ICS                & Temperature (T)                                                                                                                                               & 38                                                            \\ \midrule
NDS                & \begin{tabular}[c]{@{}l@{}}Electrical Conductivity (EC)\\ Level (L)\\ PH-Value (PH)\\ Pressure (P)\\ Temperature (T)\\ Volume (Vo)\end{tabular}               & \begin{tabular}[c]{@{}l@{}}4\\ 2\\ 4\\ 8\\ 4\\ 2\end{tabular} \\ \midrule
TCS                & \begin{tabular}[c]{@{}l@{}}Pressure (P)\\ Relative Humidity (RH)\\ Temperature (T)\\ Valve (Va)\end{tabular}                                                  & \begin{tabular}[c]{@{}l@{}}3\\ 2\\ 12\\ 3\end{tabular}        \\ \bottomrule
\end{tabular}
\caption{Sensor readings for each subsystem in the EDEN ISS dataset.}
\label{tab:edeniss_dataset}
\end{table}

\clearpage

\section[\appendixname~\thesection]{Supplemental Material for Results Section.}
\label{appendix:anomaly_types}
\begin{figure}[ht]
    \centering
    \includegraphics[width=.44\textwidth]{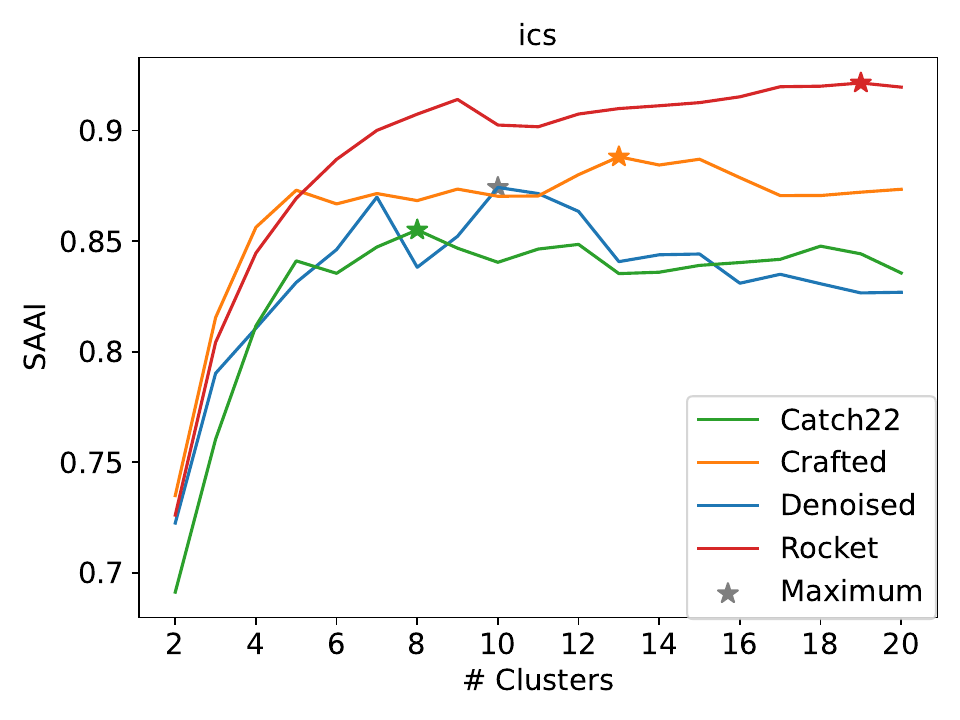}
    \caption{SAAI per feature set for K-Means clustering.}
    \label{fig:results:rq4_ics_saai}
\end{figure}

\begin{table}[ht]
  \centering
    \begin{tabular}{@{}ll@{}}
    \toprule
    \textbf{\#} & \textbf{Description}                             \\ \midrule
    0           & anomalous night phase (too cold, wrong duration) \\
    1           & peak (long)                                      \\
    2           & peak (short)                                     \\
    3           & near flat noisy or flat signal                   \\
    4           & missing / delayed warmup                         \\
    5           & anomalous day phase                              \\
    6           & interrupted cooldown / missing day-night patters \\
    7           & n/a                                              \\
    8           & flat and drop                                    \\
    9           & missing drop / missing night                     \\ \bottomrule
    \end{tabular}%
    \caption{Description of the univariate anomaly type candidates, isolated from sequence cluster plots. The numbers correspond to those given in Figure \ref{fig:results:rq4_ics_kopt_hr} and Figure \ref{fig:results:rq4_ics_k19} in the Supplemental Material and Figure 5 in the main text.}
    \label{tab:rq4_uv}
\end{table}

\clearpage

\begin{figure}[ht]
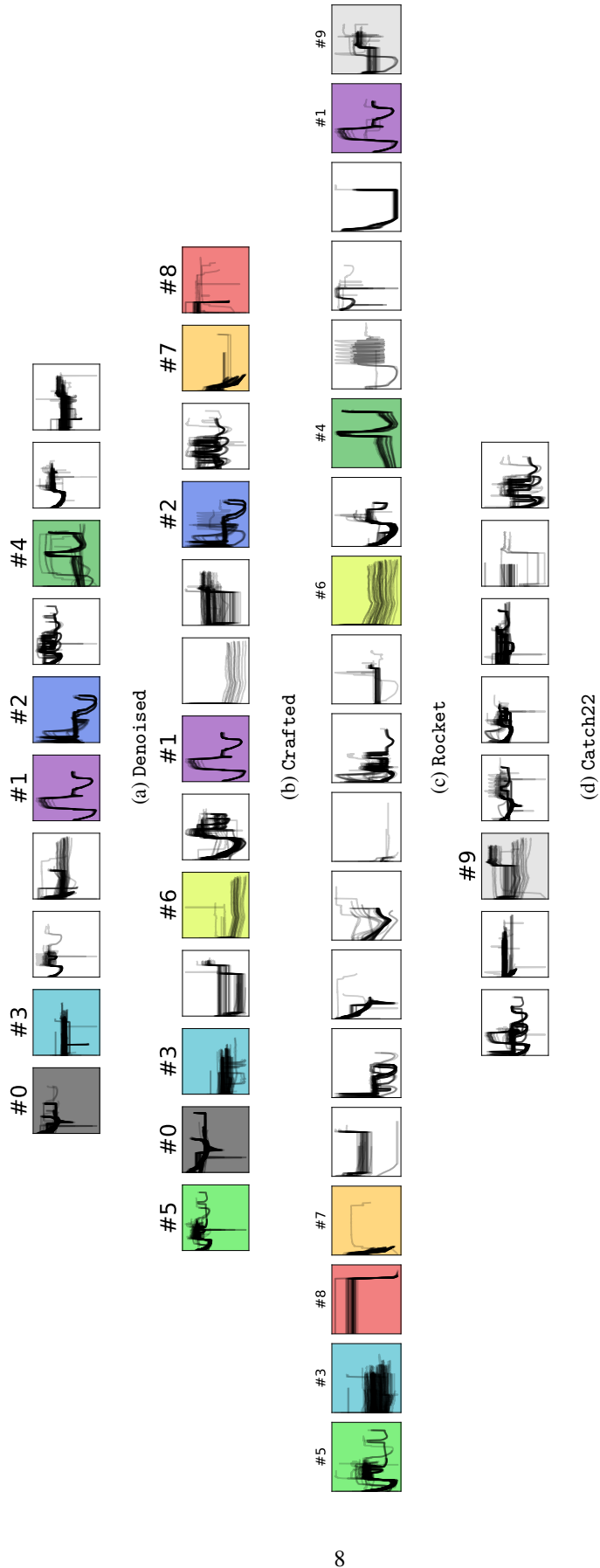

    \centering
    \rotatebox{90}{
    \begin{minipage}{\textheight}
        \centering
        \begin{subfigure}[b]{.524\textwidth}
            \centering
            \includegraphics[width=\textwidth]{img/ics/ics_sequence_plots_Denoised_k10.pdf}
            \caption{\texttt{Denoised}}
            \label{fig:ics_denoised_10_hr}
        \end{subfigure}
        \\
        \begin{subfigure}[b]{.68\textwidth}
            \centering
            \includegraphics[width=\textwidth]{img/ics/ics_sequence_plots_Crafted_k13.pdf}
            \caption{\texttt{Crafted}}
            \label{fig:ics_crafted_13_hr}
        \end{subfigure}
        \\
        \begin{subfigure}[b]{\textwidth}
            \centering
            \includegraphics[width=\textwidth]{img/ics/ics_sequence_plots_Rocket_k19.pdf}
            \caption{\texttt{Rocket}}
            \label{fig:ics_rocket_19_hr}
        \end{subfigure}
        \\
        \begin{subfigure}[b]{.42\textwidth}
            \centering
            \includegraphics[width=\textwidth]{img/ics/ics_sequence_plots_Catch22_k8.pdf}
            \caption{\texttt{Catch22}}
            \label{fig:ics_catch22_8_hr}
        \end{subfigure}
    \caption{Anomaly Clusters found in ICS temperature readings with K-Means clustering for SAAI-optimal number of clusters $K$ per feature set. Anomaly clusters consistent across different feature sets are marked by the same color. The numbers above the annotated clusters correspond to Table \ref{tab:rq4_uv}.}
    \label{fig:results:rq4_ics_kopt_hr}
    \end{minipage}%
    }
\end{figure}

\clearpage

\begin{figure}
    \centering
    \begin{subfigure}[b]{.4\textwidth}
        \includegraphics[width=\textwidth]{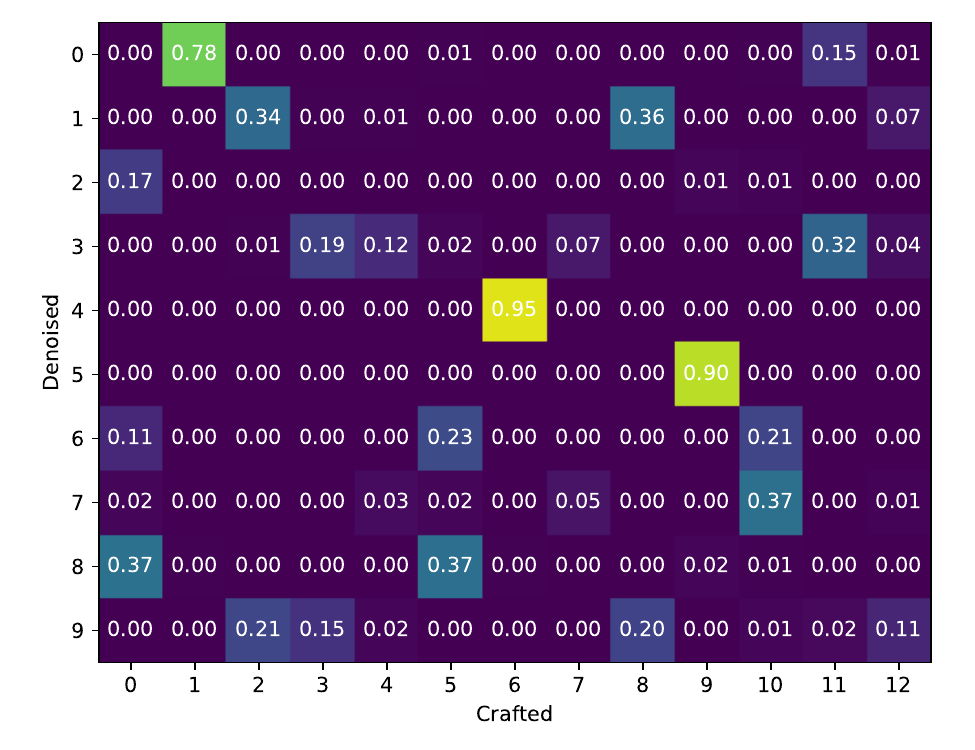}
        \caption{\texttt{Denoised} and \texttt{Crafted}}
    \end{subfigure}
    \hfill
    \begin{subfigure}[b]{.58\textwidth}
        \includegraphics[width=\textwidth]{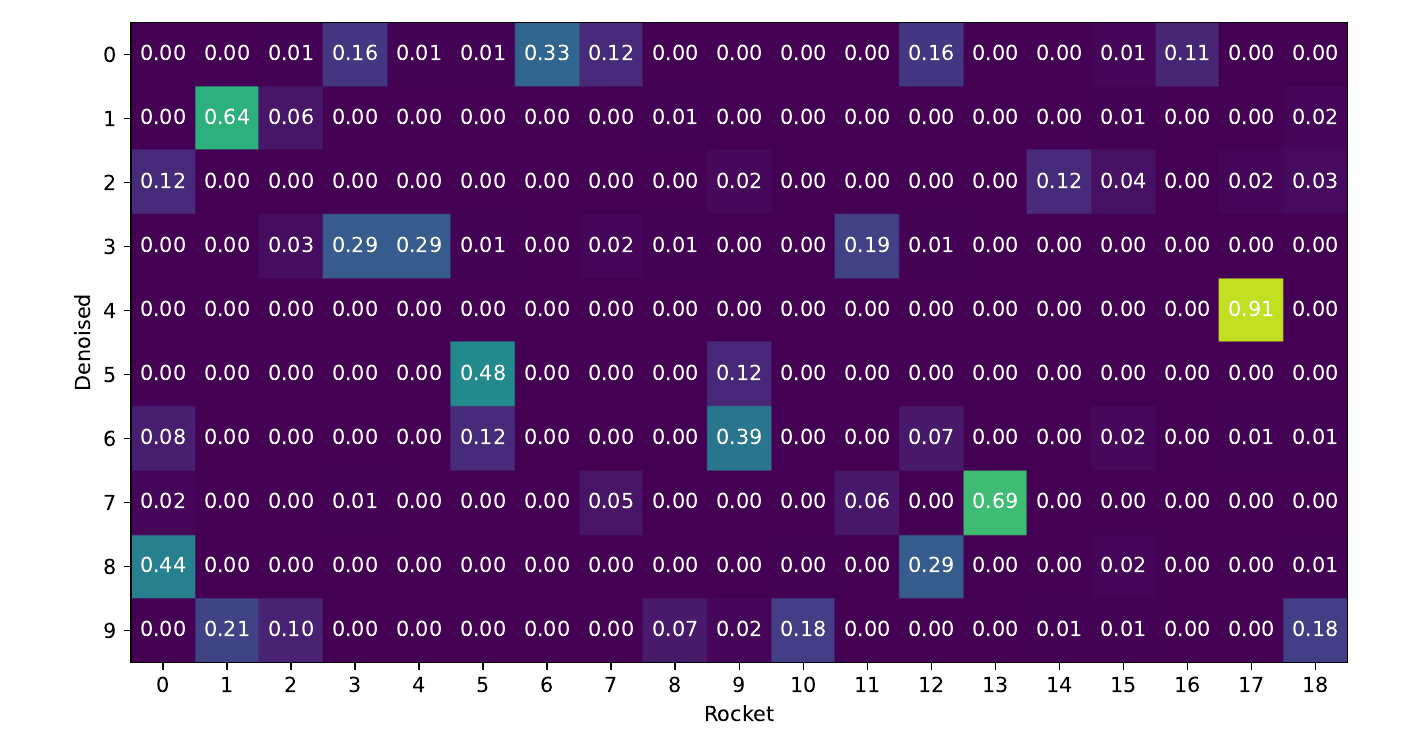}
        \caption{\texttt{Denoised} and \texttt{Rocket}}
    \end{subfigure}
    \hfill
    \begin{subfigure}[b]{.355\textwidth}
        \includegraphics[width=\textwidth]{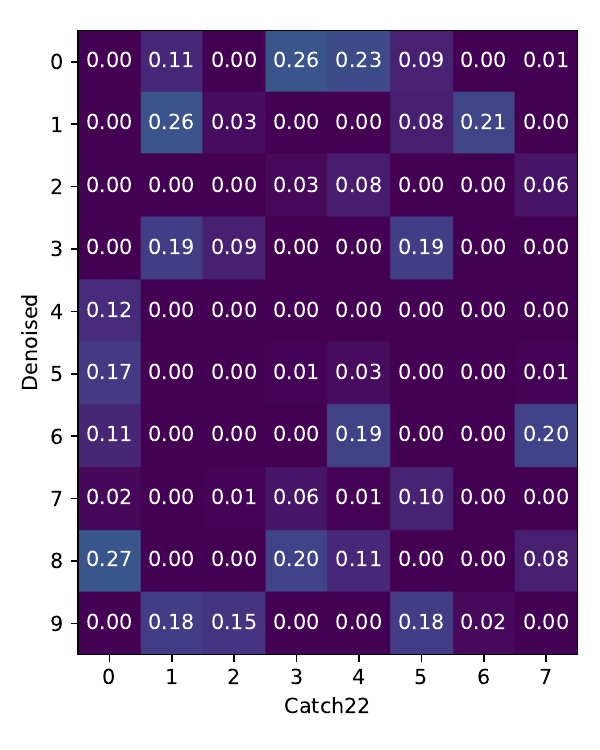}
        \caption{\texttt{Denoised} and \texttt{Catch22}}
    \end{subfigure}
    \hfill
    \begin{subfigure}[b]{.625\textwidth}
        \includegraphics[width=\textwidth]{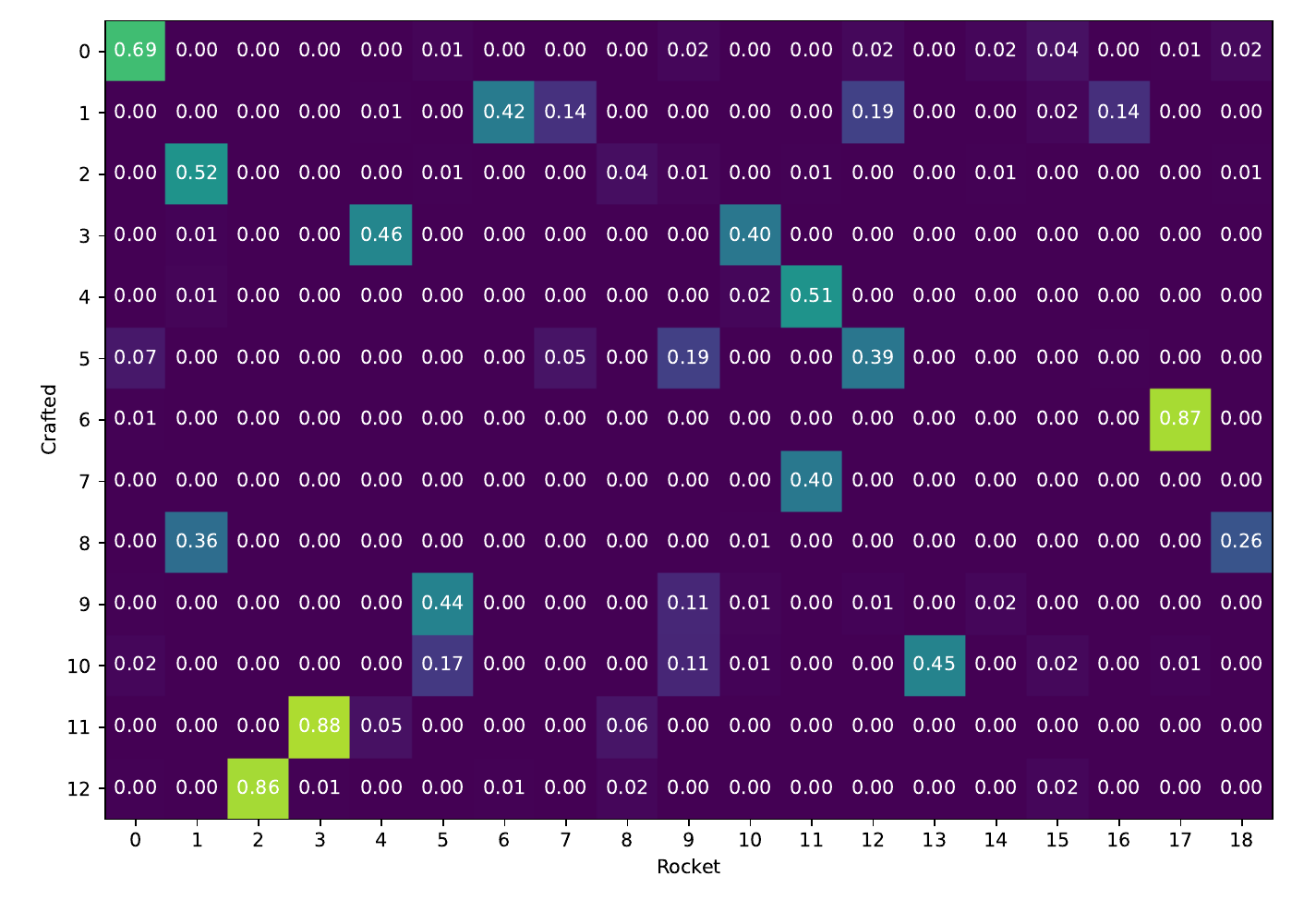}
        \caption{\texttt{Crafted} and \texttt{Rocket}}
    \end{subfigure}        
    \hfill
    \begin{subfigure}[b]{.3\textwidth}
        \includegraphics[width=\textwidth]{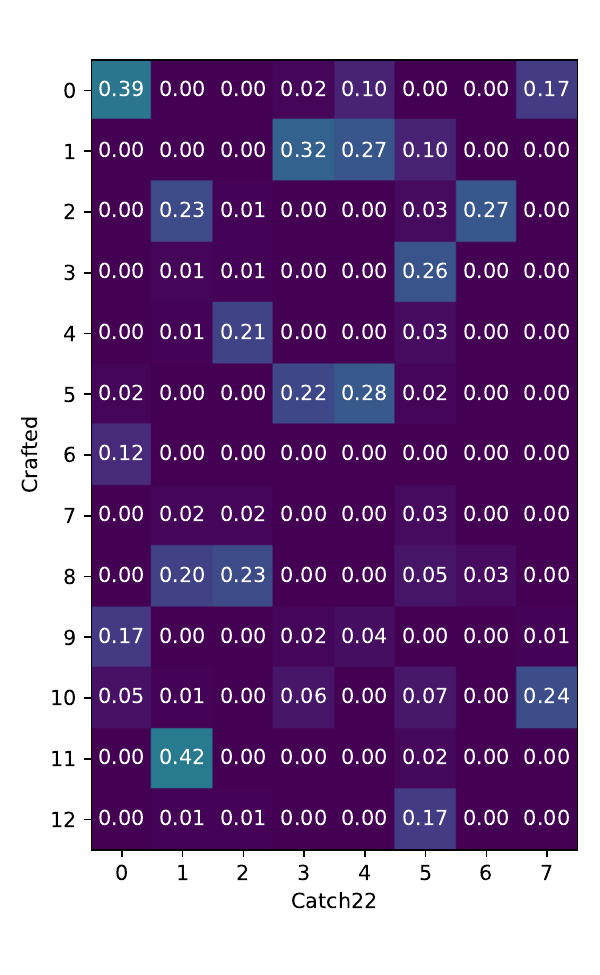}
        \caption{\texttt{Crafted} and \texttt{Catch22}}
    \end{subfigure}
    \hfill
    \begin{subfigure}[b]{.2\textwidth}
        \includegraphics[width=\textwidth]{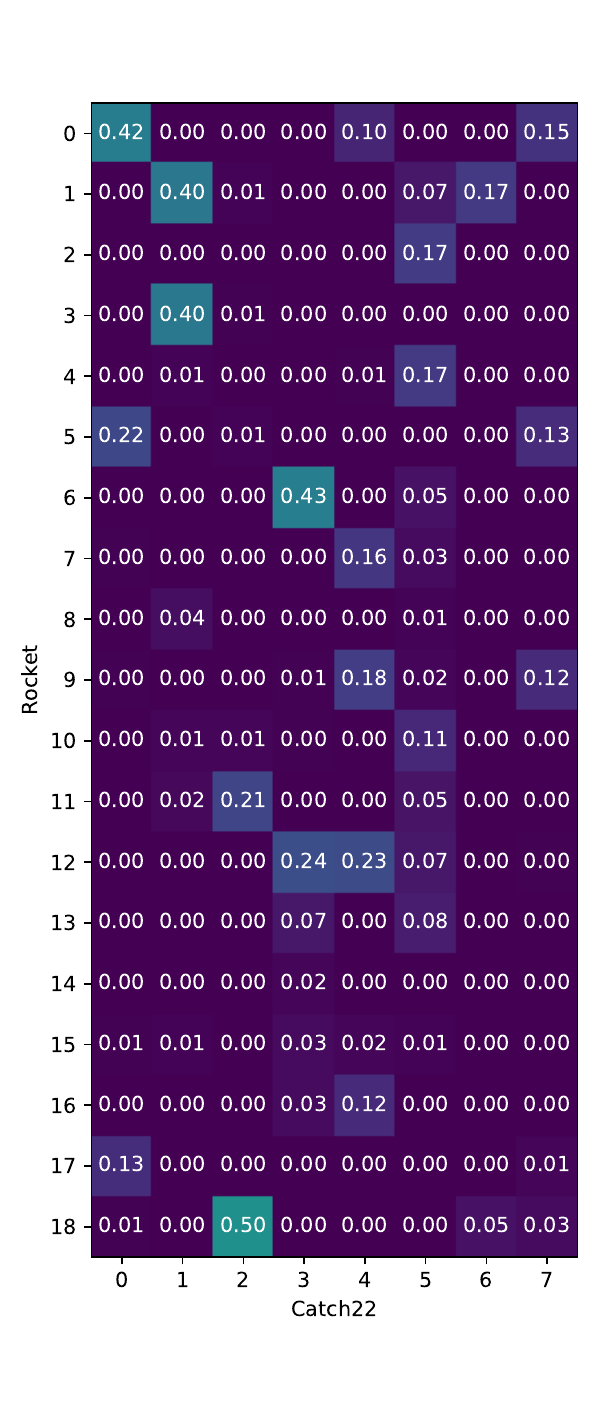}
        \caption{\texttt{Rocket} and \texttt{Catch22}}
    \end{subfigure}
    \caption{Intersection over Union of univariate ICS anomaly clusters between the four feature sets for K-Means clustering with SAAI-optimal number of clusters $K$.}
    \label{fig:ics_opt_consensus}
\end{figure}

\clearpage

\begin{figure}
    \centering
    \rotatebox{90}{
    \begin{minipage}{\textheight}
    \centering
        \begin{subfigure}[b]{\textwidth}
            \centering
            \includegraphics[width=\textwidth]{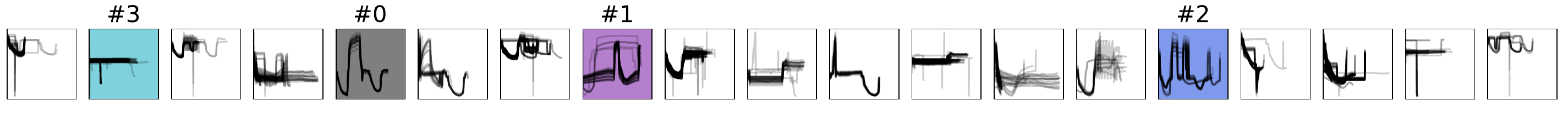}
            \caption{\texttt{Denoised}}
            \label{fig:ics_denoised_19}
        \end{subfigure}
        \hfill
        \begin{subfigure}[b]{\textwidth}
            \centering
            \includegraphics[width=\textwidth]{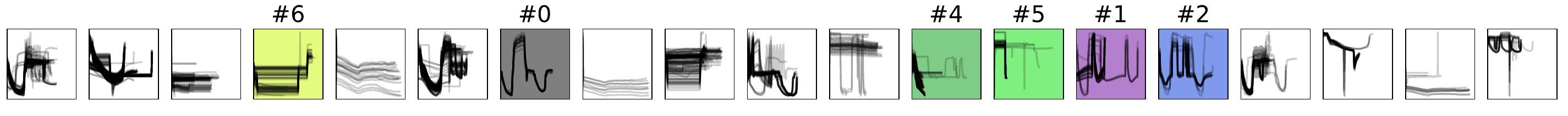}
            \caption{\texttt{Crafted}}
            \label{fig:ics_crafted_19}
        \end{subfigure}
        \hfill
        \begin{subfigure}[b]{\textwidth}
            \centering
            \includegraphics[width=\textwidth]{img/ics/ics_sequence_plots_Rocket_k19.pdf}
            \caption{\texttt{Rocket}}
            \label{fig:ics_rocket_19}
        \end{subfigure}
        \begin{subfigure}[b]{\textwidth}
            \centering
            \includegraphics[width=\textwidth]{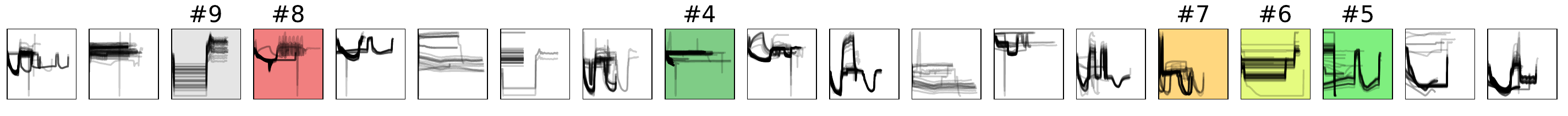}
            \caption{\texttt{Catch22}}
            \label{fig:ics_catch22_19}
        \end{subfigure}
        \caption{Anomaly Clusters found in ICS temperature readings with K-Means clustering and $K=19$. Anomaly clusters consistent across different feature sets are marked by the same color.}
        \label{fig:results:rq4_ics_k19}
    \end{minipage}%
    }
\end{figure}

\begin{figure}
    \centering
    \begin{subfigure}[b]{.42\textwidth}
        \includegraphics[width=\textwidth]{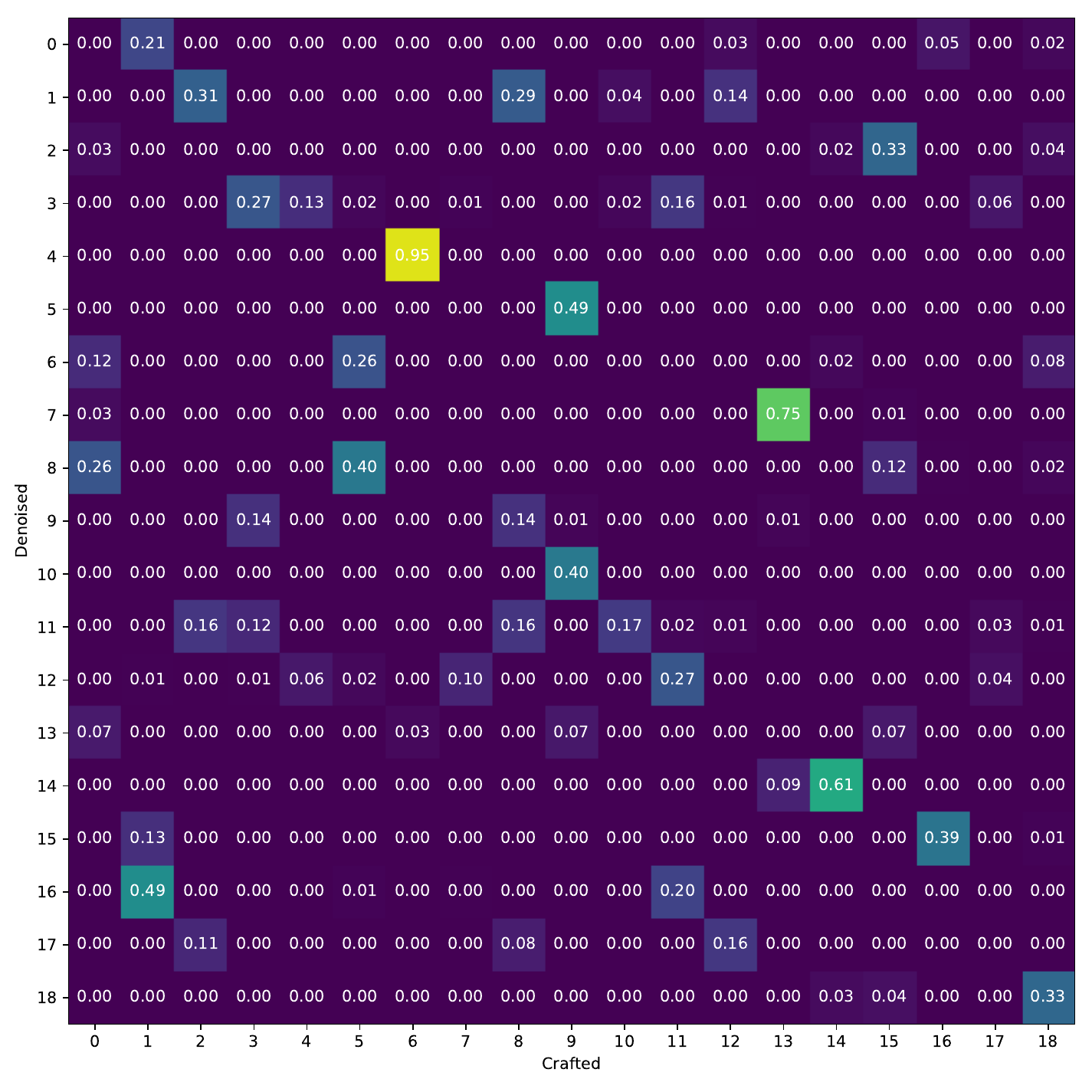}
        \caption{\texttt{Denoised} and \texttt{Crafted}}
    \end{subfigure}
    \hfill
    \begin{subfigure}[b]{.42\textwidth}
        \includegraphics[width=\textwidth]{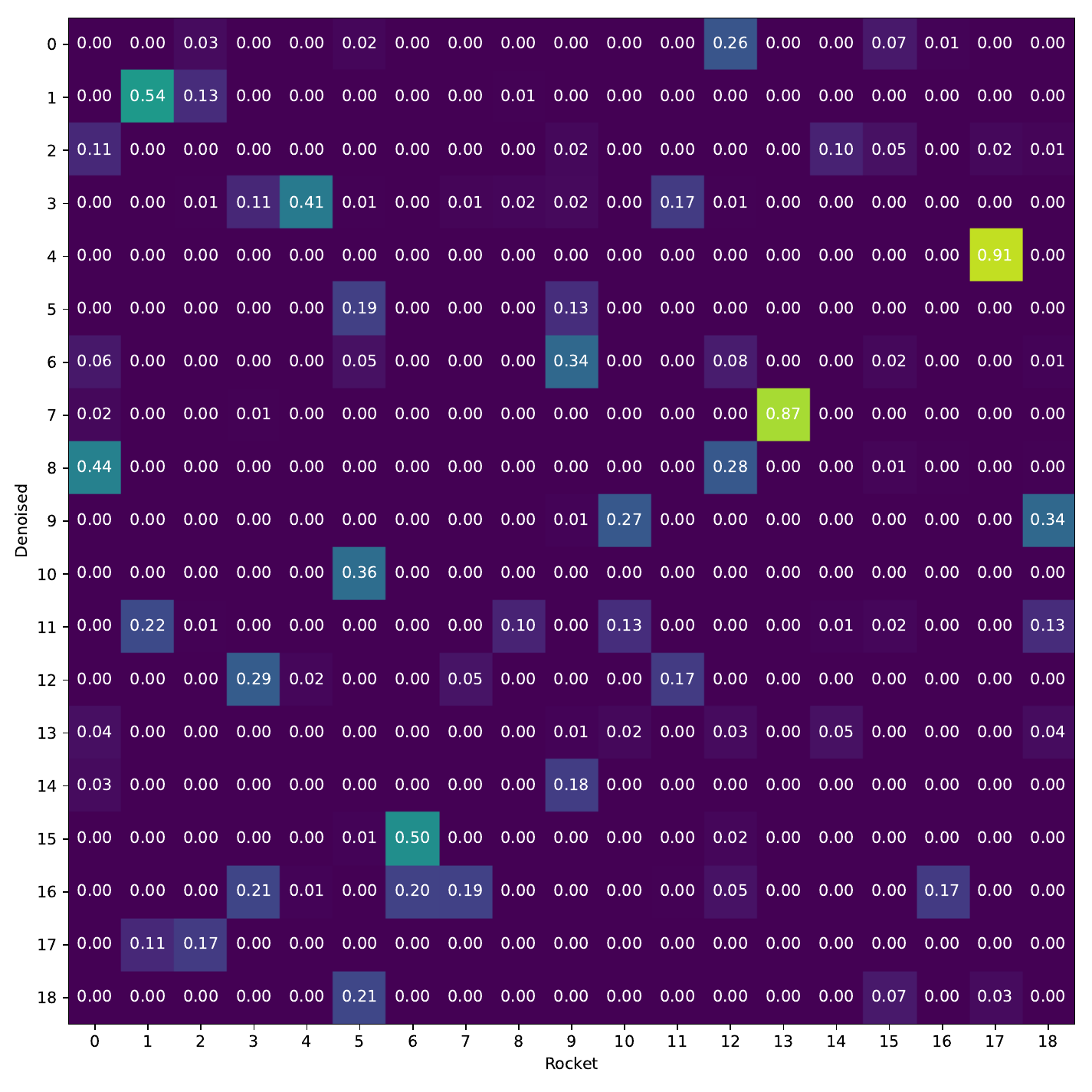}
        \caption{\texttt{Denoised} and \texttt{Rocket}}
    \end{subfigure}
    \hfill
    \begin{subfigure}[b]{.42\textwidth}
        \includegraphics[width=\textwidth]{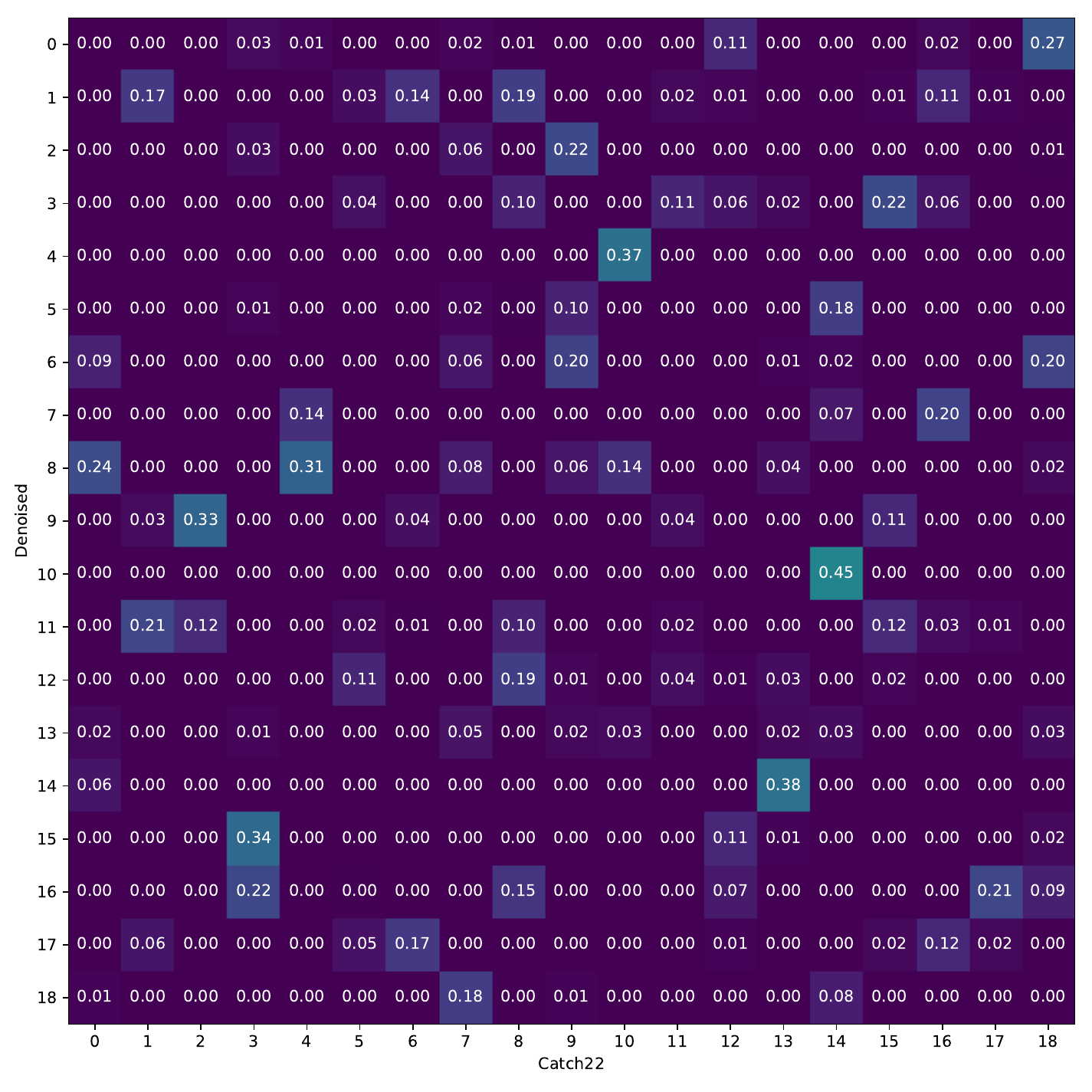}
        \caption{\texttt{Denoised} and \texttt{Catch22}}
    \end{subfigure}
    \hfill
    \begin{subfigure}[b]{.42\textwidth}
        \includegraphics[width=\textwidth]{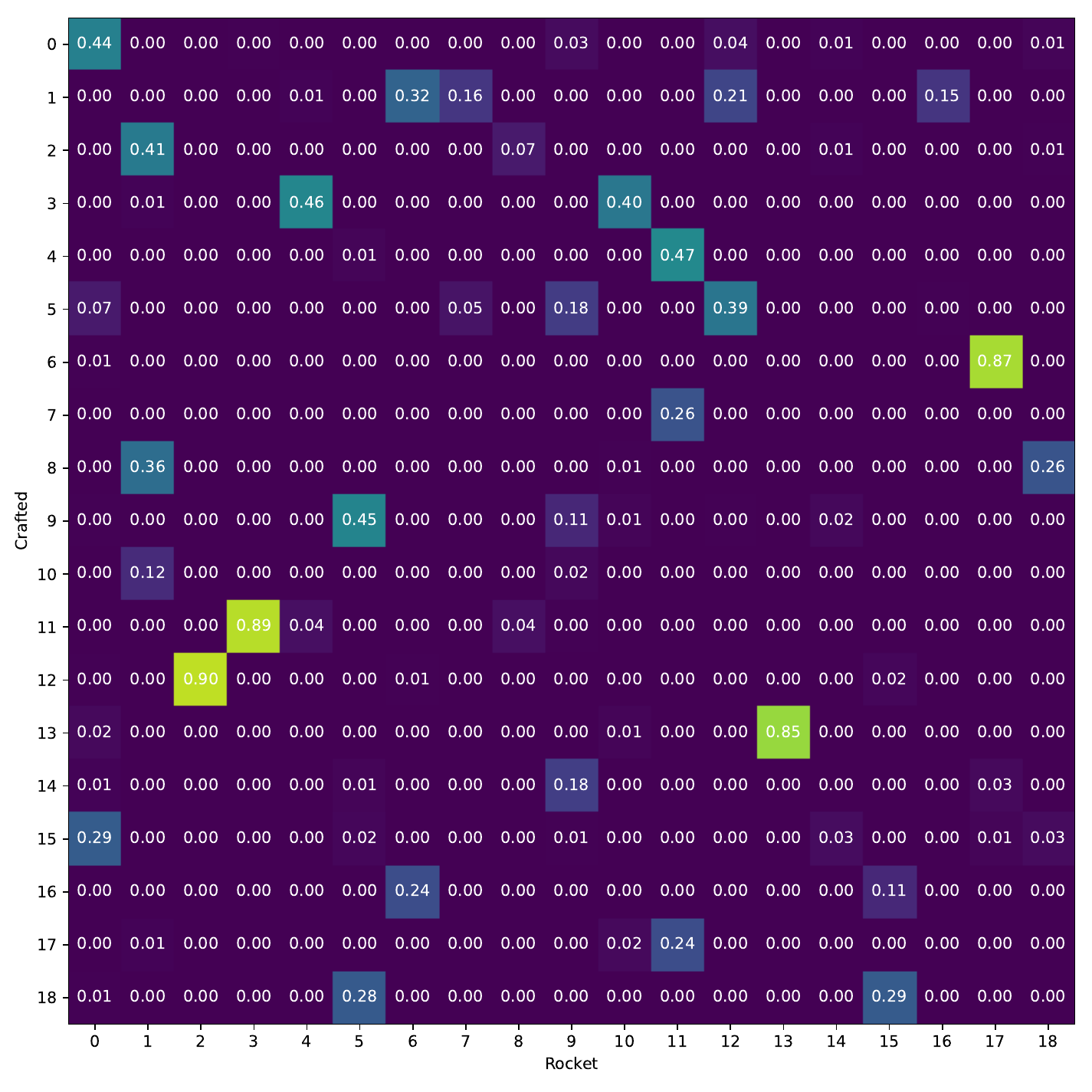}
        \caption{\texttt{Crafted} and \texttt{Rocket}}
    \end{subfigure}        
    \hfill
    \begin{subfigure}[b]{.42\textwidth}
        \includegraphics[width=\textwidth]{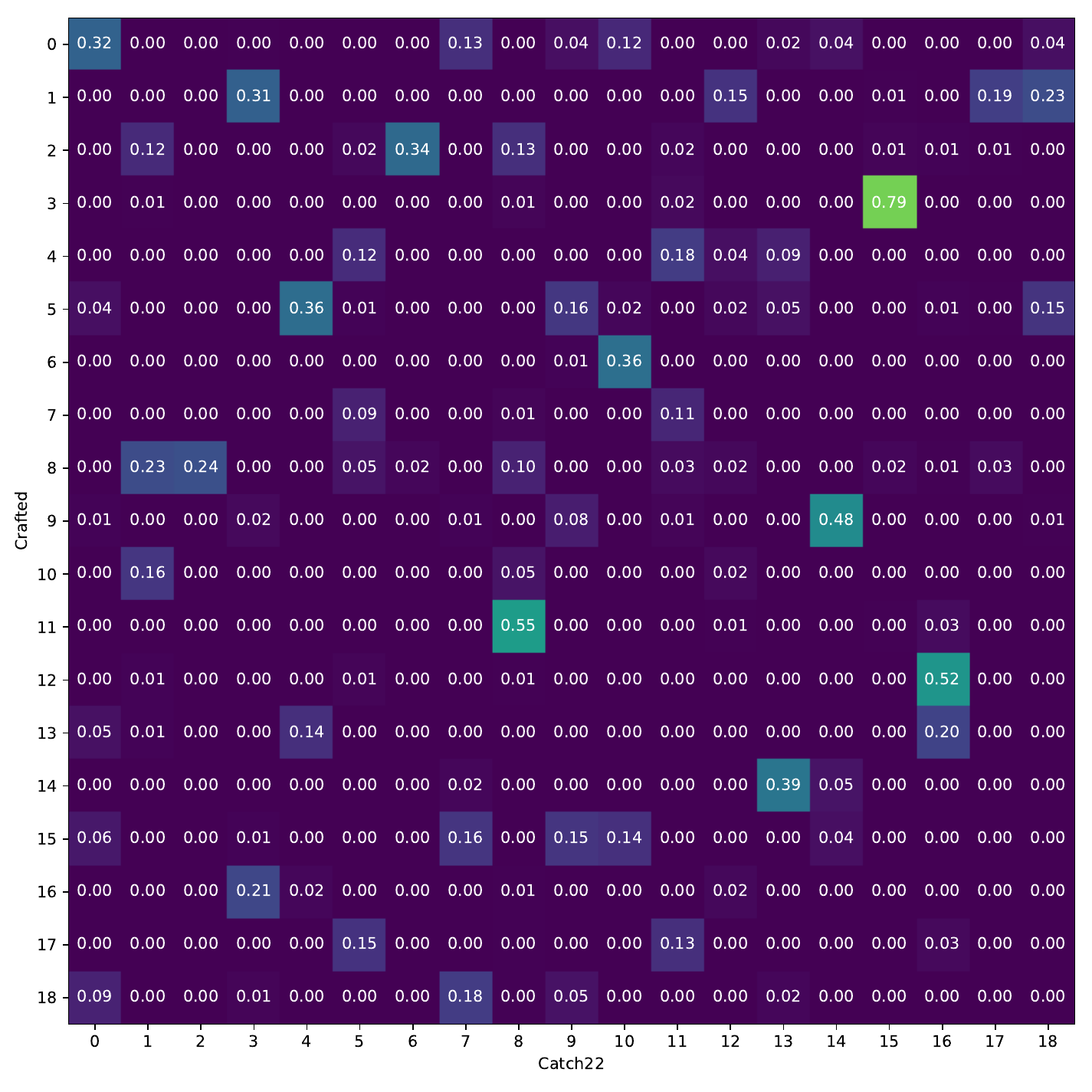}
        \caption{\texttt{Crafted} and \texttt{Catch22}}
    \end{subfigure}
    \hfill
    \begin{subfigure}[b]{.42\textwidth}
        \includegraphics[width=\textwidth]{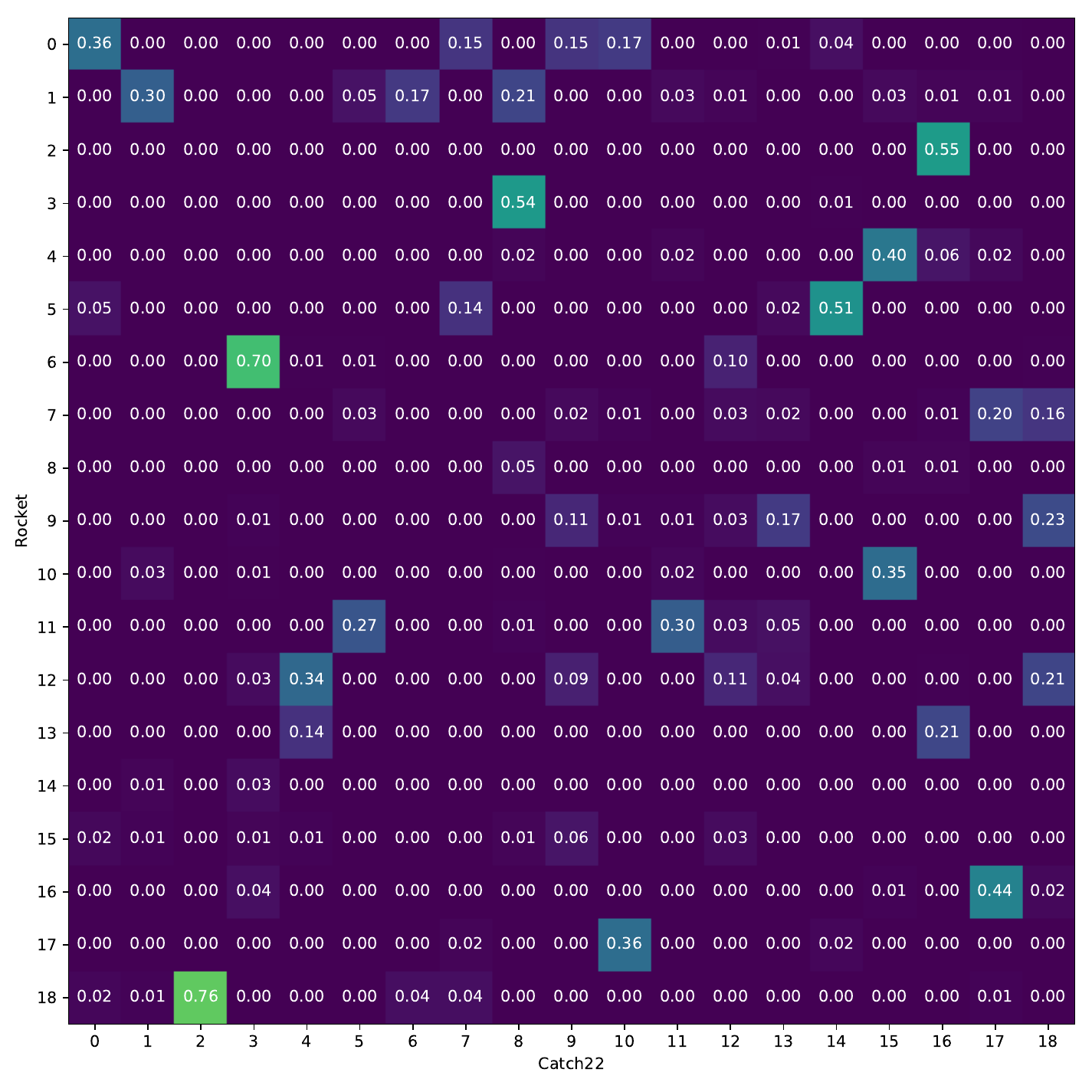}
        \caption{\texttt{Rocket} and \texttt{Catch22}}
    \end{subfigure}
    \caption{Intersection over Union of univariate ICS anomaly clusters between the four feature sets for K-Means clustering with $K=19$.}
    \label{fig:ics_k19_consensus}
\end{figure}

\clearpage

\begin{figure}[ht]
        \centering
        \begin{subfigure}[b]{\textwidth}
            \centering
            \includegraphics[width=\textwidth]{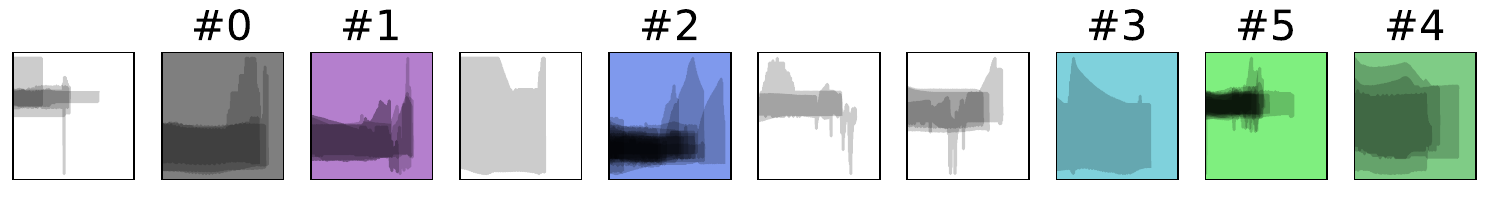}
            \caption{\texttt{Denoised}}
            \label{fig:ams-ses_denoised_10_hr}
        \end{subfigure}\\
        \begin{subfigure}[b]{\textwidth}
            \centering
            \includegraphics[width=\textwidth]{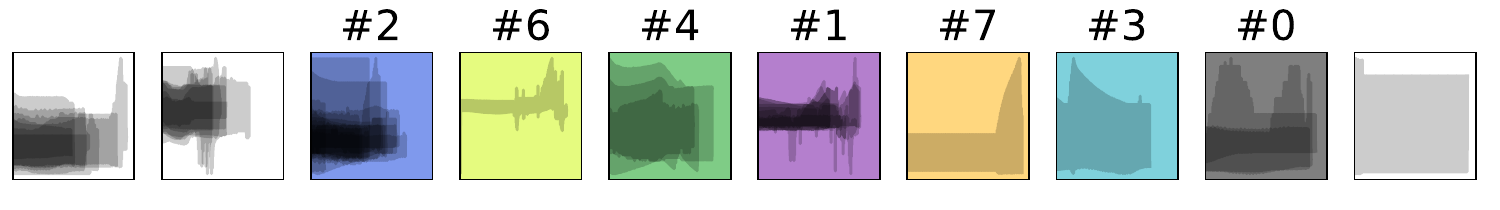}
            \caption{\texttt{Crafted}}
            \label{fig:ams-ses_crafted_10_hr}
        \end{subfigure}\\
        \begin{subfigure}[b]{\textwidth}
            \centering
            \includegraphics[width=\textwidth]{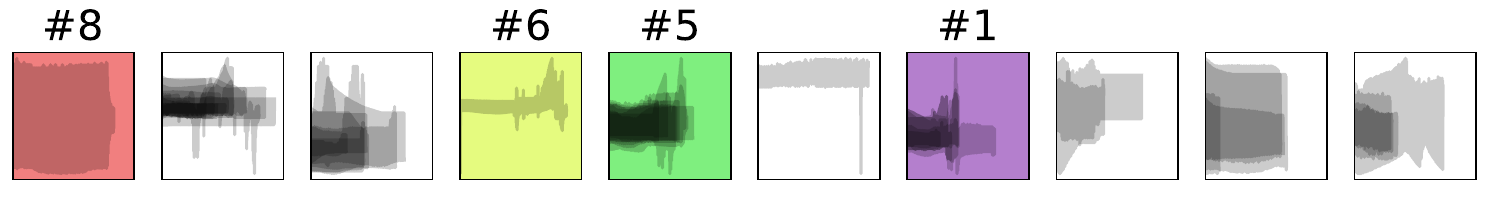}
            \caption{\texttt{Rocket}}
            \label{fig:ams-ses_rocket_10_hr}
        \end{subfigure}\\
        \begin{subfigure}[b]{\textwidth}
            \centering
            \includegraphics[width=\textwidth]{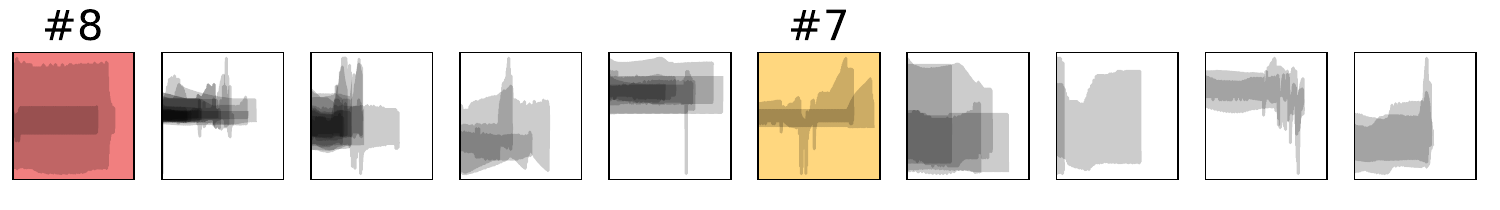}
            \caption{\texttt{Catch22}}
            \label{fig:ams-ses_catch22_10_hr}
        \end{subfigure}
        \caption{Anomaly Clusters found in NDS nutrient solution level measurements with K-Means clustering for $K=7$. Anomaly Clusters that are consistent accross different feature sets are marked by the same color.}
        \label{fig:results:rq4_ams-ses_k10_hr}
\end{figure}

\clearpage

\begin{figure}
    \centering
    \begin{subfigure}[b]{.42\textwidth}
        \includegraphics[width=\textwidth]{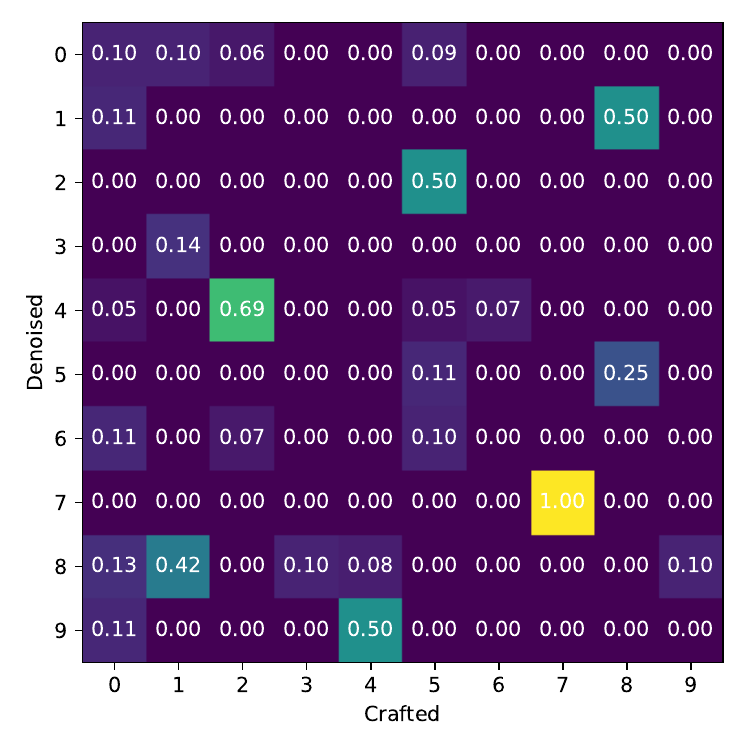}
        \caption{\texttt{Denoised} and \texttt{Crafted}}
    \end{subfigure}
    \hfill
    \begin{subfigure}[b]{.42\textwidth}
        \includegraphics[width=\textwidth]{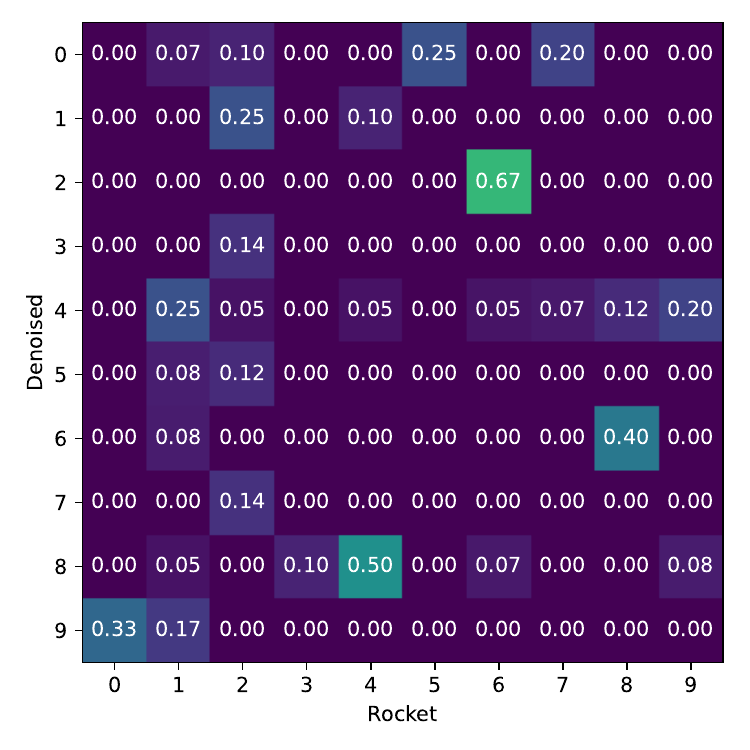}
        \caption{\texttt{Denoised} and \texttt{Rocket}}
    \end{subfigure}
    \hfill
    \begin{subfigure}[b]{.42\textwidth}
        \includegraphics[width=\textwidth]{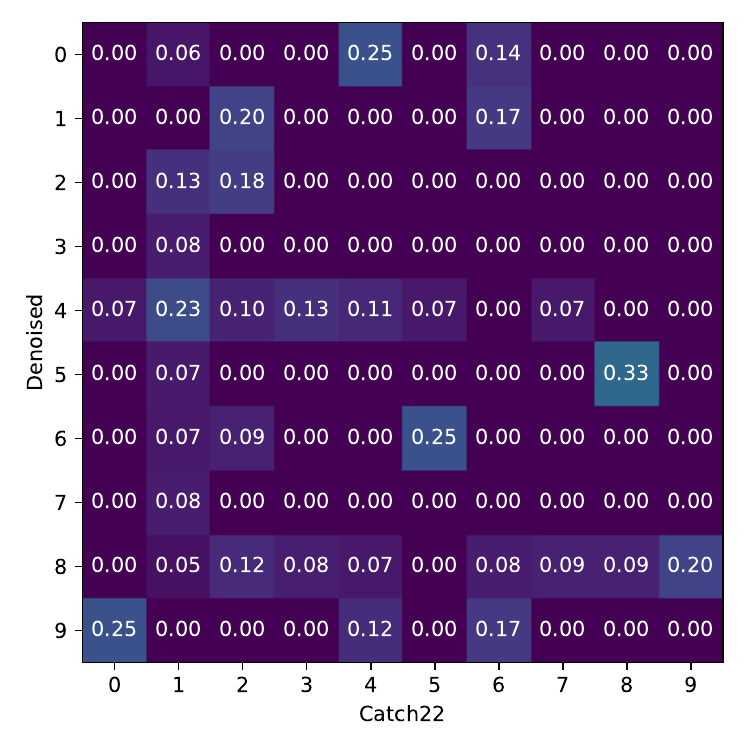}
        \caption{\texttt{Denoised} and \texttt{Catch22}}
    \end{subfigure}
    \hfill
    \begin{subfigure}[b]{.42\textwidth}
        \includegraphics[width=\textwidth]{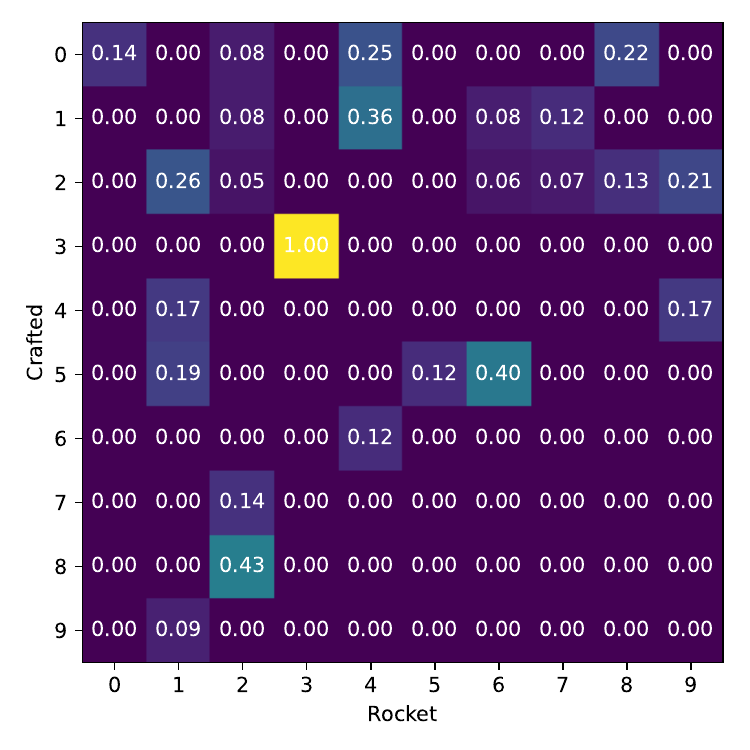}
        \caption{\texttt{Crafted} and \texttt{Rocket}}
    \end{subfigure}        
    \hfill
    \begin{subfigure}[b]{.42\textwidth}
        \includegraphics[width=\textwidth]{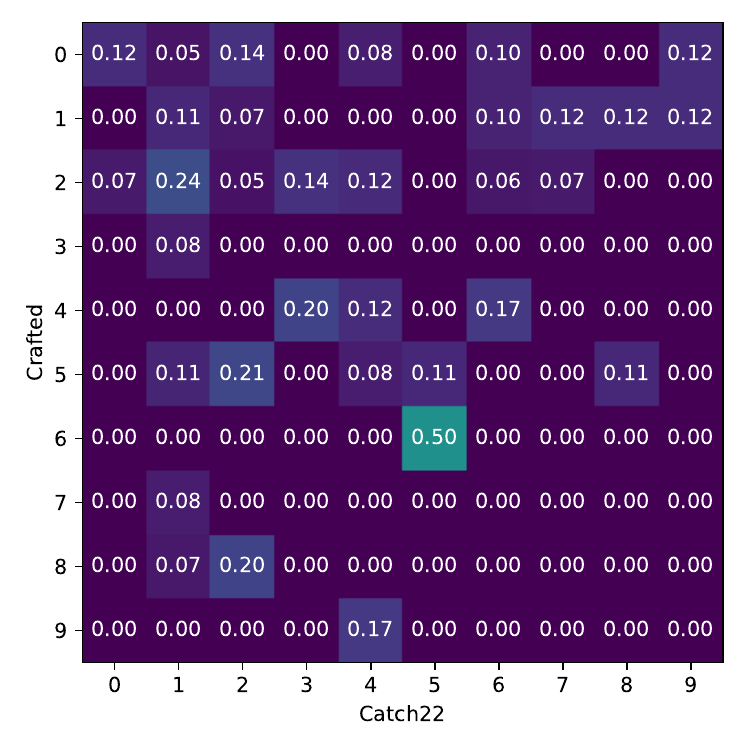}
        \caption{\texttt{Crafted} and \texttt{Catch22}}
    \end{subfigure}
    \hfill
    \begin{subfigure}[b]{.42\textwidth}
        \includegraphics[width=\textwidth]{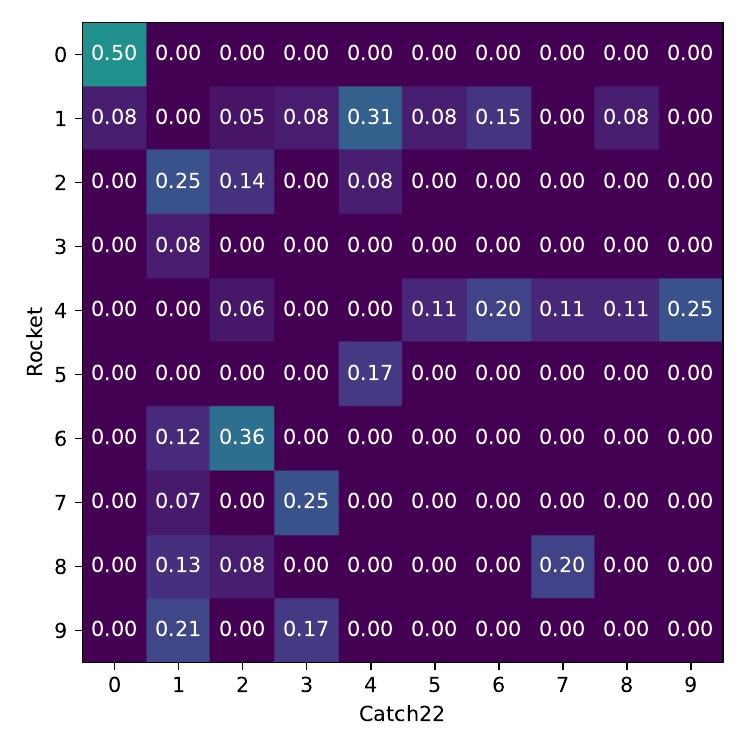}
        \caption{\texttt{Rocket} and \texttt{Catch22}}
    \end{subfigure}
    \caption{Intersection over Union of multivariate AMS-SES anomaly clusters between the four feature sets for K-Means clustering with $K=10$.}
    \label{fig:amsses_consensus}
\end{figure}

\clearpage

\begin{table}[ht]
    \centering
    \begin{tabular}{@{}ll@{}}
    \toprule
    \textbf{\#} & \textbf{Description}          \\ \midrule
    0           & n/a                           \\
    1           & $CO^2$ Peak                      \\
    2           & n/a                           \\
    3           & n/a                           \\
    4           & Temperature+$CO^2$ Peak  with RH Drop    \\
    5           & Level Shift                   \\
    6           & Steep increase at startup     \\
    7           & Temperature (air-in) extremum \\
    8           & Steep increase / decrease      \\ \bottomrule
    \end{tabular}%
    \caption{Description of the multivariate anomaly type candidates, isolated from the AMS-SES readings. The numbers correspond to those given in Figure 19 in the  Supplemental Material and Figure \ref{fig:results:rq4_ams-ses_k10_hr} in the main text.}
    \label{tab:rq4_mv}
\end{table}

\begin{figure}[ht]
    \centering
    \includegraphics[width=.95\textwidth]{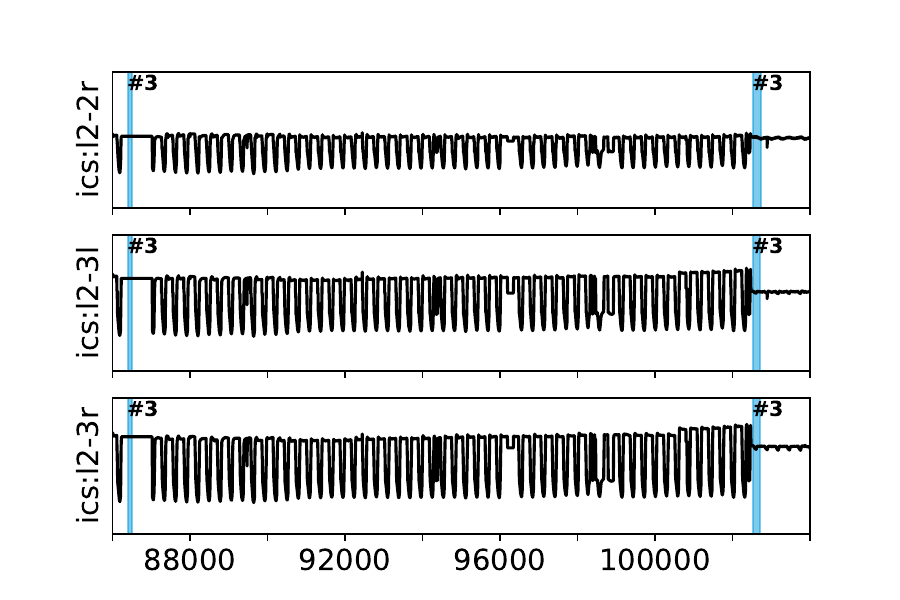}
    \caption{Candidates for recurring univariate anomalous behavior: "near flat or flat signal" (\#3)) anomalies highlighted in a subset of the ICS temperature readings as detected by the \texttt{Crafted} features.}
    \label{fig:rq5:ics:3_hq}
\end{figure}

\begin{figure}[ht]
    \centering
    \includegraphics[width=.95\textwidth]{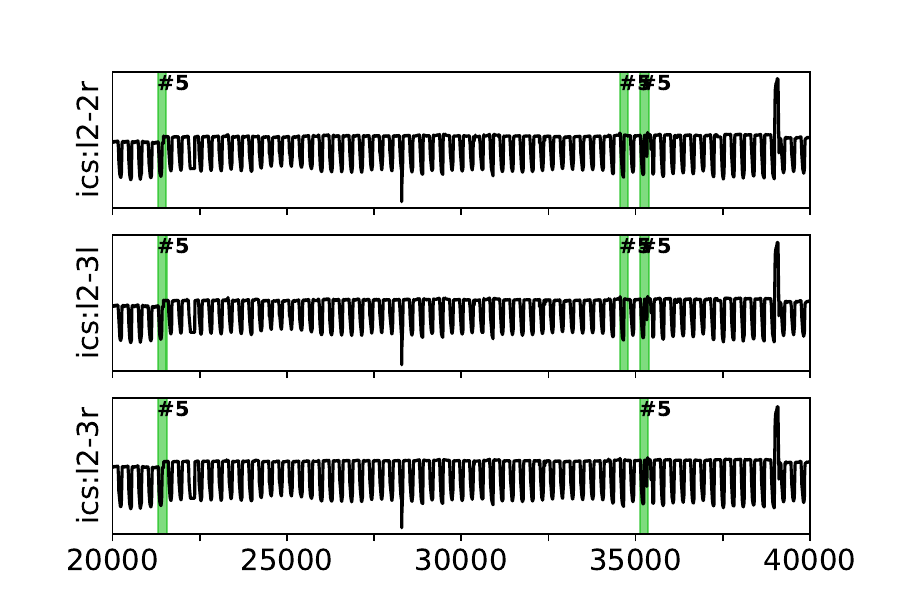}
    \caption{Candidates for recurring univariate anomalous behavior: "anomalous day phase" (\#5)) anomalies highlighted in a subset of the ICS temperature readings as detected by the \texttt{Rocket} features.}
    \label{fig:rq5:ics:5_hq}
\end{figure}

\begin{figure}[ht]
    \begin{subfigure}[b]{0.49\textwidth}
        \centering
        \includegraphics[width=\textwidth]{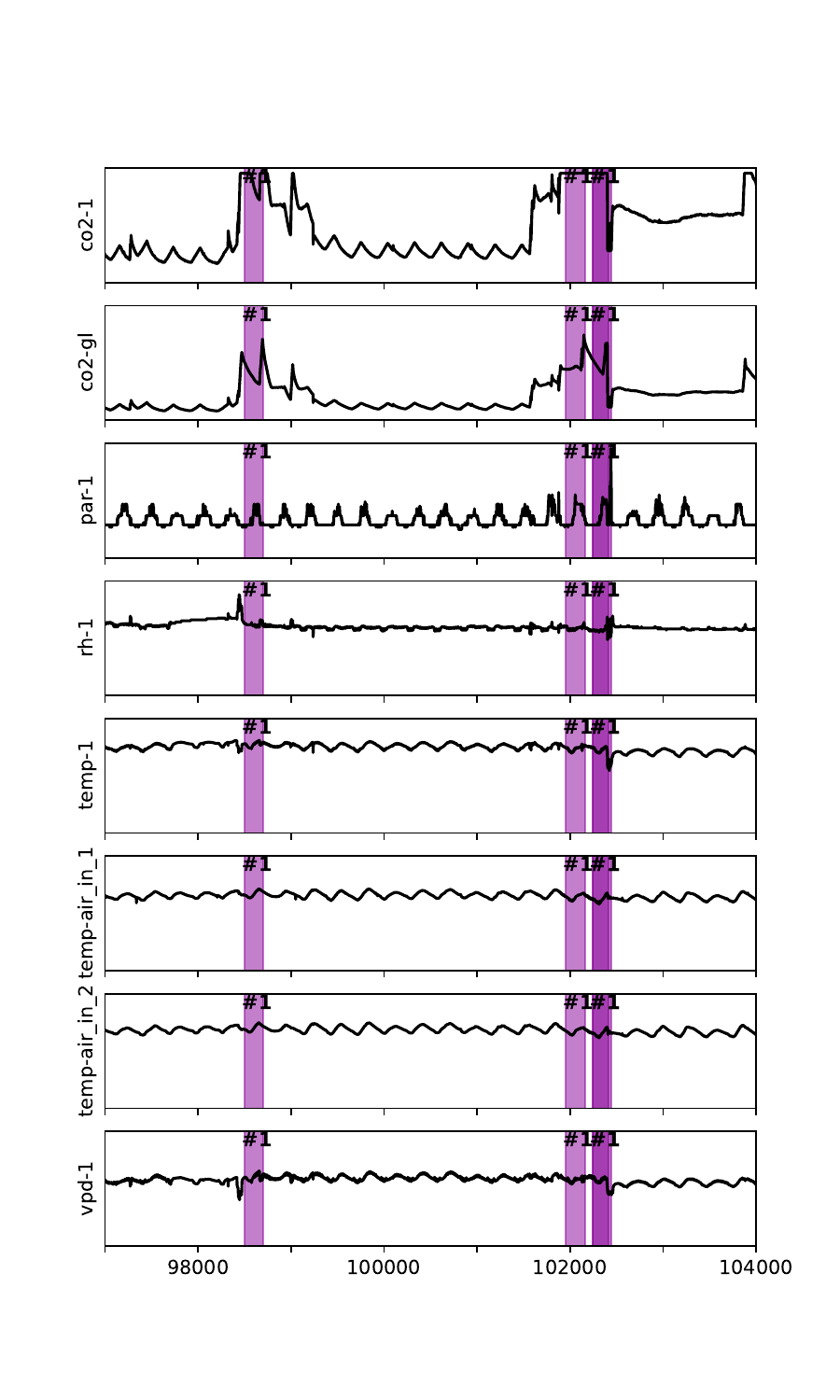}
        \caption{}
        \label{fig:rq5:ams-ses:1_hq}
    \end{subfigure}
    \begin{subfigure}[b]{0.49\textwidth}
        \centering
        \includegraphics[width=\textwidth]{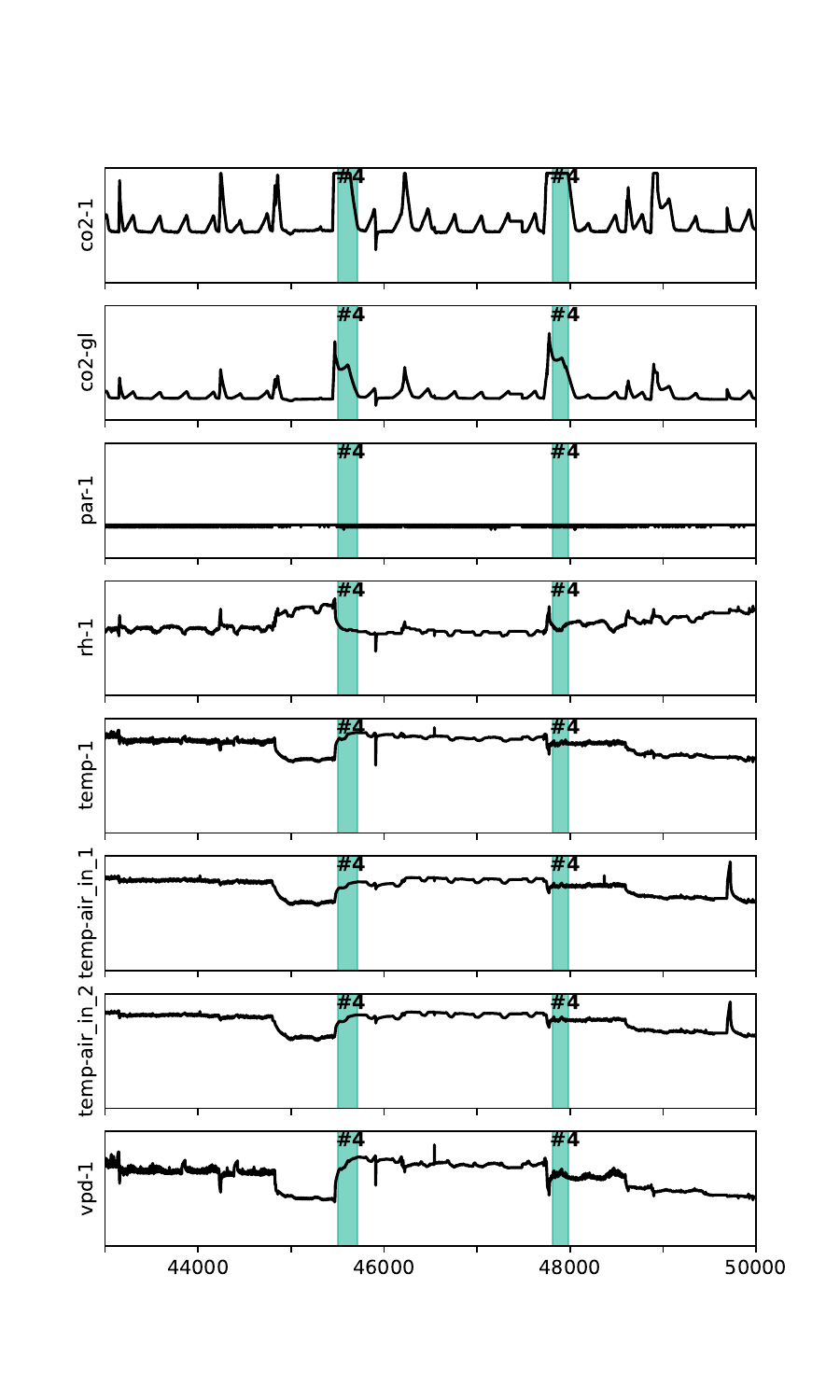}
        \caption{}
        \label{fig:rq5:ams-ses:4_hq}
    \end{subfigure}
    \caption{Candidates for recurring multivariate anomalous behavior: (a) "$CO^2$ peak" (\#1)) and (b) "temperature and $CO^2$ peak with relative humidity drop" anomalies highlighted in the AMS-SES readings as detected by the \texttt{Crafted} features.}
    \label{fig:results:rq5:hq}
\end{figure}

\end{document}